\documentclass[letterpaper]{article} 
\usepackage{aaai24}  
\usepackage{times}  
\usepackage{helvet}  
\usepackage{courier}  
\usepackage[hyphens]{url}  
\usepackage{graphicx} 
\usepackage{natbib}  
\usepackage{caption} 
\usepackage{algorithm}
\usepackage{algorithmic}
\usepackage{amsmath,amsfonts}
\usepackage{algorithmic}
\usepackage{algorithm}
\usepackage{array}
\usepackage{textcomp}
\usepackage{url}
\usepackage{verbatim}
\usepackage{graphicx}
\usepackage{multirow}
\usepackage{newfloat}
\usepackage{listings}
\DeclareCaptionStyle{ruled}{labelfont=normalfont,labelsep=colon,strut=off} 
\floatstyle{ruled}
\newfloat{listing}{tb}{lst}{}
\floatname{listing}{Listing}
\title{Voxel or Pillar: Exploring Efficient Point Cloud Representation \\ 
for 3D Object Detection}
\author{
    Yuhao Huang, Sanping Zhou, Junjie Zhang, Jinpeng Dong, Nanning Zheng\thanks{Corresponding author.}
}
\affiliations{
    National Key Laboratory of Human-Machine Hybrid Augmented Intelligence,\\
    National Engineering Research Center for Visual Information and Applications,\\
    and Institute of Artificial Intelligence and Robotics, Xi'an Jiaotong University\\
     \{hyh950623@stu, spzhou@mail, hooz1009@stu, djp1235a@stu, nnzheng@mail\}.xjtu.edu.cn
}

\begin{document}

\maketitle

\begin{abstract}
Efficient representation of point clouds is fundamental for LiDAR-based 3D object detection. While recent grid-based detectors often encode point clouds into either voxels or pillars, the distinctions between these approaches remain underexplored. In this paper, we quantify the differences between the current encoding paradigms and highlight the limited vertical learning within. To tackle these limitations, we introduce a hybrid Voxel-Pillar Fusion network~(VPF), which synergistically combines the unique strengths of both voxels and pillars. Specifically, we first develop a sparse voxel-pillar encoder that encodes point clouds into voxel and pillar features through 3D and 2D sparse convolutions respectively, and then introduce the Sparse Fusion Layer~(SFL), facilitating bidirectional interaction between sparse voxel and pillar features. Our efficient, fully sparse method can be seamlessly integrated into both dense and sparse detectors. Leveraging this powerful yet straightforward framework, VPF delivers competitive performance, achieving real-time inference speeds on the nuScenes and Waymo Open Dataset. The code will be available at \url{https://github.com/HuangYuhao-0623/VPF}.
\end{abstract}

\section{Introduction}
\label{sec:intro}

LiDAR-based 3D object detection methods have been widely adopted in autonomous driving and robot navigation systems, as point clouds from LiDAR sensors reflect geometric information explicitly and are rarely affected by weather conditions. Unlike 2D images, point clouds are sparse and non-uniformly distributed, which poses two challenges for 3D object detection: constructing efficient detection network and establishing robust object representation from points with varying distributions. 

\begin{figure}[t!]
  \centering
  \small
  \begin{minipage}[t]{0.95\linewidth}
    \centering
    \includegraphics[width=\linewidth]{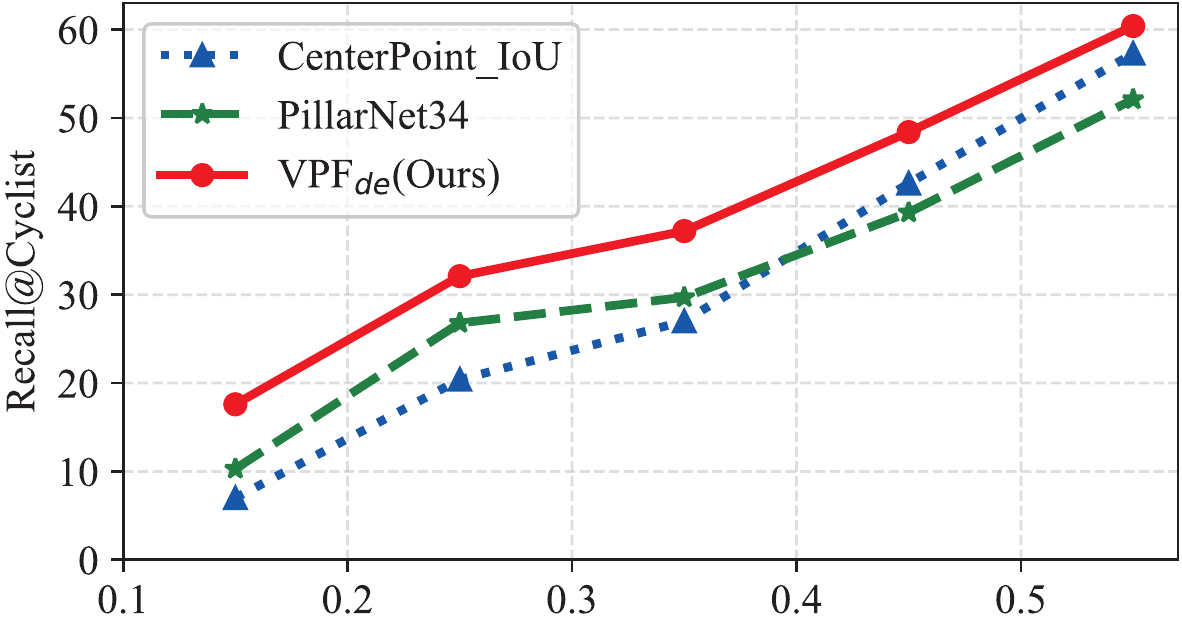}
  \end{minipage}

  \vspace{3pt}

  \begin{minipage}[t]{0.95\linewidth}
    \centering
    \includegraphics[width=\linewidth]{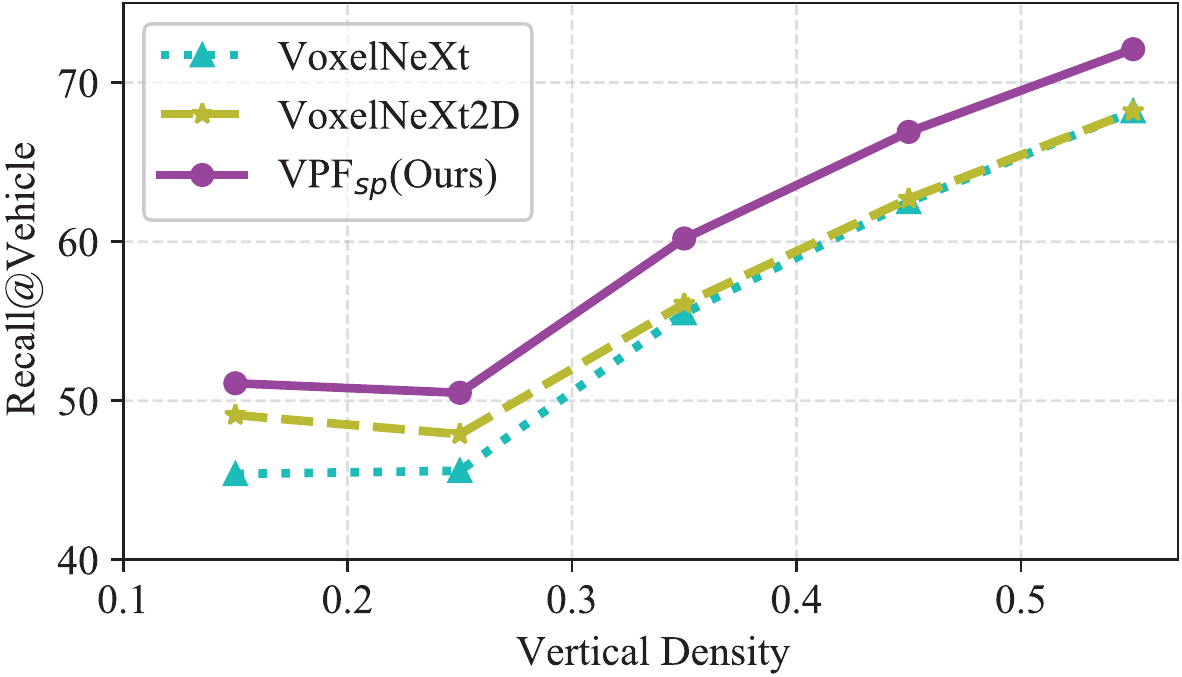}
  \end{minipage}
  \caption{Recall vs. Vertical Density comparison. For both dense and sparse detectors~\cite{yin2021center,shi2022pillarnet,chen2023voxelnext}, pillar-based representations show enhanced recall under low vertical densities, while voxel-based representations tend to excel in high-density scenarios. Notably, our hybrid representation offers consistent improvements across different situations.}
  \label{fig:vpf_statistics}
\end{figure}

There are several paradigms for point cloud representations. Range-view-based methods~\cite{GregoryPMeyer2019LaserNetAE, AlexBewley2020RangeCD, Fan2021RangeDetID} convert point clouds into compact 2.5D range images and apply well-studied 2D detectors~\cite{Ren2015FasterRT,Lin2017FeaturePN} to predict 3D boxes. While efficient, these methods may distort the geometry of 3D point clouds, and introduce the scale variance problems~\cite{YanghaoLi2019ScaleAwareTN}. Alternatively, point-based detectors~\cite{CharlesRQi2019DeepHV,ShaoshuaiShi2018PointRCNN3O,ZetongYang20203DSSDP3,YifanZhang2022NotAP} extract point-wise features with PointNet series directly~\cite{Qi2017PointNetDL, Qi2017PointNetDH}, and benefit from flexible receptive fields. Nevertheless, they suffer from the time-consuming spherical query and aggressive downsampling strategy. Finally, grid-based methods~\cite{yin2021center, hu2022afdetv2, shi2022pillarnet,shi2020pv,He2022VoxelST} quantize the point cloud into regular voxels or pillars and generate bounding boxes in 2D Bird's Eye View (BEV), achieving notable balance in both performance and efficiency.

While both the voxel and pillar are prevalent representations in grid-based methods, their distinctions are rarely discussed. Given their different behavior in the vertical direction, we conduct an exploratory experiment on Waymo Open Dataset~\cite{sun2020scalability} to analyze the consequential effects. We first split the ground truth (GT) according to its vertical point distribution: Given each GT box, we uniformly divide it into $10$ bins in the vertical direction, then calculate the vertical density as $S_{Z}=N_{bin}/10$, where $N_{bin}$ denotes the number of non-empty bins. Next, we count the recall vs. vertical density curve of several grid-based detectors\footnote{We adopt single-stage voxel-encoded CenterPoint with the same detection head as PillarNet, as detailed in Ablation Studies.}~\cite{yin2021center, shi2022pillarnet, chen2023voxelnext}. As illustrated in Fig.~\ref{fig:vpf_statistics}, Voxel-based detectors (i.e., CenterPoint\_IoU and VoxelNeXt) show consistent better performance when the vertical density is high, while pillar-based methods (i.e., PillarNet and VoxelNeXt2D) excel in situations with low density. 

This observation underscores the limited vertical representation in the current voxel and pillar-based methods. Specifically, detectors utilizing voxels typically employ the 3D submanifold convolution~\cite{Graham2017SubmanifoldSC}, which confines feature diffusion from non-empty voxels to empty ones, thereby restricting the receptive field. On the contrary, pillar-based methods discretized point clouds into vertical volumes, each of which encodes all its neighboring points in a certain X-Y coordinate. While this paradigm obtains the full-ranged vertical receptive field, it encounters challenges in capturing fine-grained features and is prone to significant information loss, particularly when processing areas with high point density. Given the limitations of both voxels and pillars, we are inspired to develop a method that \textbf{synergistically combines them for a more robust representation, especially in the vertical direction, while maintaining computational efficiency.}

In this paper, we propose Voxel-Pillar Fusion (VPF), a hybrid point cloud representation designed to synergistically harness the strengths of both voxels and pillars. We begin by crafting a sparse voxel-pillar encoder that segments point clouds into voxels and pillars, subsequently encoding these sparse volume features through 3D and 2D sparse convolutions. To enlarge the vertical receptive field of voxels and enrich the fine-grained information in pillars, we present the Sparse Fusion Layer (SFL) to establish the voxel-pillar bidirectional interaction. Specifically, SFL aggregates voxel features vertically and broadcasts pillar features to their corresponding vertical voxels. It then integrates these aggregated and broadcasted features with the original pillar and voxel features. Moreover, our method, both computationally efficient and fully sparse, can be seamlessly incorporated into both dense and sparse detectors.~\cite{yin2021center, shi2022pillarnet, chen2023voxelnext}. We also conduct comprehensive experiments on nuScenes~\cite{caesar2020nuscenes} and Waymo Open Dataset~\cite{sun2020scalability}. The results indicate that our method achieves state-of-the-art performance with real-time inference speed, and enhances the vertical representation from both voxels and pillars as showcased in Fig.~\ref{fig:vpf_statistics}.

Our contributions are summarized as follows:
\begin{itemize}
  \item We highlight the limitations in vertical representation learning of current grid-based methods and introduce a hybrid point cloud representation that synergistically harnesses the strengths of both voxels and pillars.
  \item We propose the Sparse Fusion Layer, which facilitates the voxel-pillar bidirectional interaction to enlarge the vertical receptive field of voxels and enrich the fine-grained information in pillars.
  \item We devise both dense and sparse detectors based on our proposed hybrid representation. Through comprehensive experiments on large-scale datasets, we validate the significance and practicality of the voxel-pillar fusion.
\end{itemize}

\section{Related Work}
\label{sec:related}

\subsection{Grid-based 3D Object Detection}

Grid-based methods primarily utilize voxel or pillar representations. Voxel-based detectors divide point clouds into 3D voxels and deploy 3D convolutions for voxel-wise feature extraction. Pioneering work VoxelNet~\cite{Zhou2018VoxelNetEL} replaces the hand-crafted representation with the voxel feature encoding layer, enabling the end-to-end training procedure. Then, SECOND~\cite{YanYan2018SECONDSE} introduces the 3D sparse convolution~\cite{BenGraham2015Sparse3C, Graham2017SubmanifoldSC} to avoid the redundant computation on empty voxels. Two-stage detectors~\cite{JiajunDeng2021VoxelRT,yin2021center} adopt the coarse-to-fine pattern in 2D detection~\cite{Ren2015FasterRT, Lin2017FeaturePN}. For instance, Voxel R-CNN~\cite{JiajunDeng2021VoxelRT} proposes the voxel RoI pooling to extract RoI features from the voxels. As a contrast, single-stage methods~\cite{Zheng2021CIASSDCI, hu2022afdetv2} directly generate bounding boxes in one stage. Some recent works~\cite{He2022VoxelST,wang2023dsvt,yang2023gd} have been investigating the powerful transformer architecture~\cite{AshishVaswani2017AttentionIA, AlexeyDosovitskiy2020AnII, KirillovAlexander2020EndtoEndOD} in 3D object detection, while FSD~\cite{LueFan2022FullyS3} and VoxelNeXt~\cite{chen2023voxelnext} exploring the fully sparse detection framework. For Pillar-based methods~\cite{Yang2018PIXORR3, lang2019pointpillars, shi2022pillarnet}, point clouds are encoded into 2D volumes (pillars). PointPillars~\cite{lang2019pointpillars} converts point clouds to pillars and deploys PointNet~\cite{Qi2017PointNetDL} for pillar-wise feature extraction. PillarNet~\cite{shi2022pillarnet} proposes the 2D sparse backbone for efficient pillar encoding, achieving a favorable trade-off between performance and inference speed. Then, SST~\cite{fan2022embracing} introduces a single-stride transformer to enhance performance in the challenging pedestrian category. Though voxel or pillar representations are widely used, their differences are rarely discussed.

\begin{figure*}[t!]
  \centering
  \includegraphics[width=\textwidth]{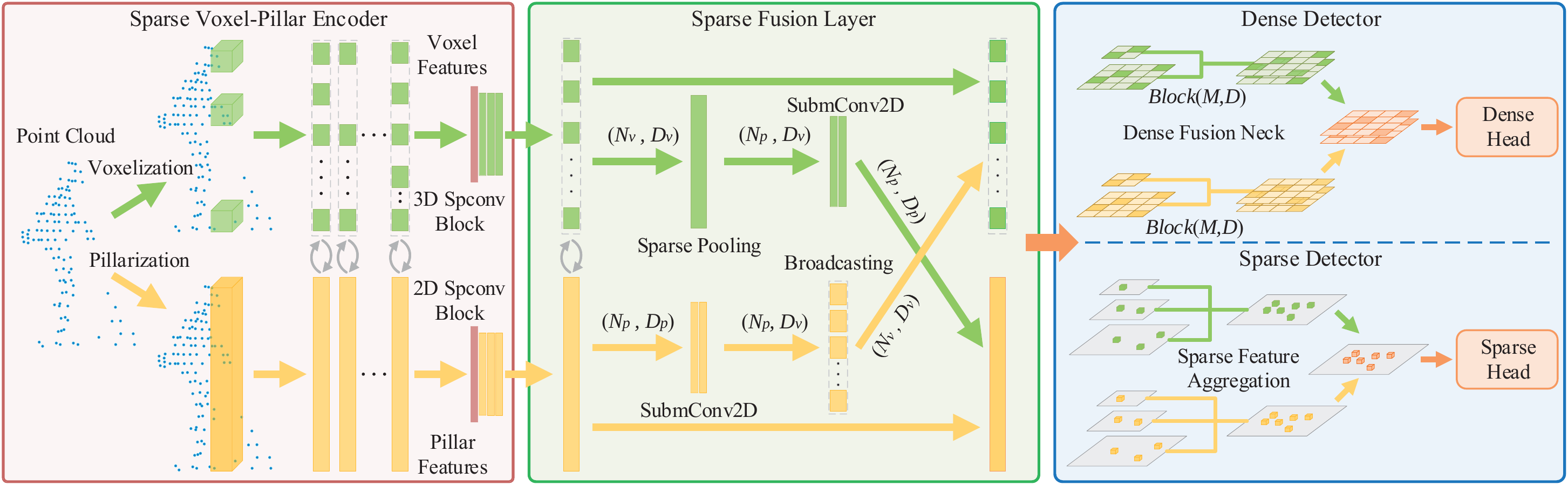}
  \caption{The framework of VPF. Point clouds are first processed by the sparse voxel-pillar encoder, which extracts correlated sparse voxel and pillar features. The subsequent Sparse Fusion Layer facilitates bidirectional interaction, capturing supplementary information from both types of sparse features. Together, these components form a hybrid backbone capable of integrating with both dense and sparse detectors.}
  \label{fig:vpfusion}
\end{figure*}

\subsection{Multi-Source Feature Fusion}

Methods of multi-source fusion~\cite{Yang2019STDS3, SuPang2020CLOCsCO, zhou2020hierarchicaledge} are proposed to combine distinct information~\cite{zhou2023inverse} from different sources. For instance, point-voxel join representation~\cite{Yang2019STDS3, shi2020points, shi2020pv} is presented to integrate the flexible receptive field with efficient feature learning schemes. Besides, multi-view fusion~\cite{XiaozhiChen2016MultiView3O,zhou2020end, wang2020pillar, xin2019semi} is another paradigm for information supplementation. MVF~\cite{zhou2020end} utilizes the complementary information from both BEV and perspective view, while Pillar-OD~\cite{wang2020pillar} applies pillar encoder in BEV and cylindrical view. Recently, multi-modal fusion~\cite{MingLiang2019MultiTaskMF,SuPang2020CLOCsCO, YingweiLi2022DeepFusionLD} has achieved remarkable progress. CLOCS~\cite{SuPang2020CLOCsCO} proposes a late fusion strategy to exploit the geometric and semantic consistencies between 2D and 3D predictions. DeepFusion~\cite{YingweiLi2022DeepFusionLD} presents InverseAug and LearnableAlign to align multi-modal features in the late stage. While multi-source fusion provides additional information from different sources, it also raises a problem. As the separate sources usually vary, effective alignment of multi-source features becomes challenging and sophisticated. In our work, we design a straightforward yet efficient structure for voxel-pillar fusion.

\section{Methodology}
\label{sec:method}

In pursuit of enhancing the vertical point cloud representation, we propose the Voxel-Pillar Fusion (VPF), a hybrid point cloud representation harnessing both 2D and 3D volume information. As shown in Fig.~\ref{fig:vpfusion}, point clouds are initially quantized into voxels and pillars with the same resolution in the X-Y plane. Then, we adopt the sparse Voxel-Pillar encoder with four intermediate steps, each of which features a sparse convolution block, with 3D and 2D sparse convolutions for voxel and pillar feature extraction, respectively. To enrich the local context for the pillar branch and infuse the vertical semantics into the voxel branch, we deploy the Sparse Fusion Layer (SFL) at the end of each step, creating a bidirectional interaction between the voxel and pillar features. Finally, we design both dense and sparse detectors, {\em i.e.}, VPF$_\mathrm{de}$ and VPF$_\mathrm{sp}$, which are equipped with our presented hybrid representation backbone.

\subsection{Sparse Voxel-Pillar Encoder}
\label{sec3:backbone}

\subsubsection{Consistent Voxel-Pillar Encoding.} 
While voxels and pillars represent point clouds differently, at the same resolution in the X-Y plane, the vertical collection of voxels contains the same point clouds as the corresponding pillar's under specific X-Y coordinates. We exploit this trait to construct a consistent voxel-pillar encoder. 

Given a point cloud $\mathcal{S}=\{ s_{i} \in \mathbb{R}^{4} \}^{N}_{i=1}$, where $N$ is the number of points. We first divide the 3D space into voxels and pillars with spatial resolution of $L \times W \times H$ and $L \times W$, respectively. Next, dynamic voxelization~\cite{zhou2020end} is deployed for initial sparse voxel and pillar feature generation. In the voxel branch, $\mathcal{S}$ is quantized based on a pre-defined voxel size. This process yields point-to-voxel indices, represented as $\mathcal{V}=\{v_i=(l_i, w_i, h_i) \in \mathbb{N}^{3} \}^{N}_{i=1}$. Subsequently, the initial voxel features are derived by computing the mean of all points in each non-empty voxel with a sparse mean-pooling~\cite{FeyLenssen2019}. In the pillar branch, due to the consistency of voxels and pillars in Bird's Eye View~(BEV), the point-to-pillar indices, $\mathcal{P}=\{p_i=(l_i, w_i) \in \mathbb{N}^{2} \}^{N}_{i=1}$ are obtained by removing the vertical indices. Finally, we apply PointNet~\cite{lang2019pointpillars} with sparse max-pooling to form the initial pillar features.

\subsubsection{Sparse Conv Block.} 
\begin{figure}[t!]
  \centering
  \includegraphics[width=\linewidth]{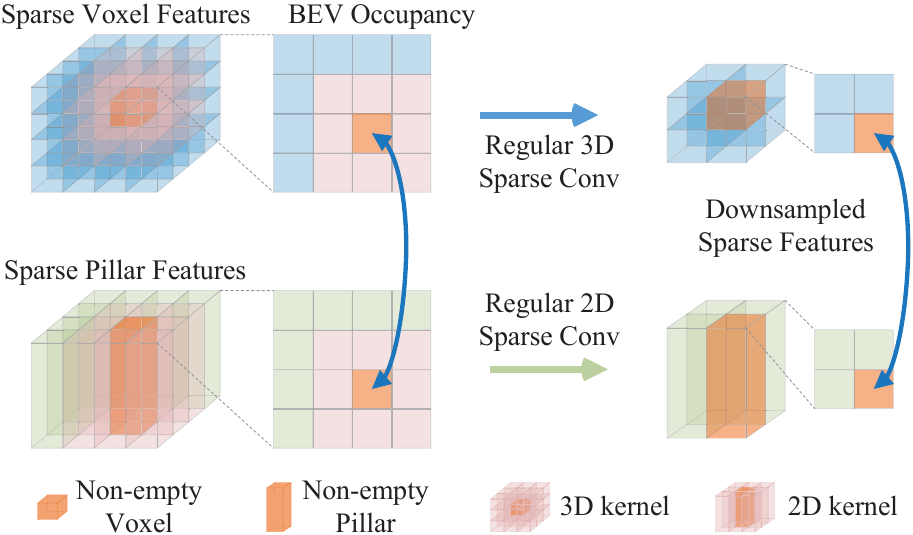}
  \caption{Consistent voxel-pillar downsampling process. In the downsampling procedure, by equalizing the kernel sizes, strides, and padding operations of 2D and 3D regular sparse convolutions in X-Y dimensions, the consistent BEV occupancy is preserved for sparse voxel and pillar features.}
  \label{fig:consist_vpencode}
\end{figure}

We extract sparse voxel and pillar features via dual-branch sparse convolution (conv) blocks, which include both 3D and 2D sparse convolutions. Similar to previous single branch framework~\cite{YanYan2018SECONDSE, shi2022pillarnet}, for each block, the voxel branch consists of a 3D regular sparse convolution followed by multiple 3D submanifold sparse convolutional layers, while the pillar branch is composed of its 2D equivalents. The regular sparse convolutions are applied for spatial downsampling, and submanifold sparse convolutions ensure that the output locations are identical to the input locations to optimize efficiency. As depicted in Fig.~\ref{fig:consist_vpencode}, for both 2D and 3D regular sparse convolutions, we equalize the kernel size, stride, and padding operations in the horizontal dimension ({\em i.e.}, $L$ and $W$ dimensions). This ensures that voxel and pillar features expand to the same X-Y plane location. In such manner, non-empty voxels and pillars possess consistent occupancy in BEV, {\em i.e.}, each non-empty pillar has several corresponding non-empty voxels ($\ge 1$) at the same X-Y coordinates. This consistency facilitates the fusion of sparse voxel and pillar features.

Overall, the sparse voxel-pillar encoder consists of 4 intermediate steps, where sparse conv blocks are deployed to sequentially generate sparse voxel and pillar features with $1\times,~2\times,~4\times,$ and $8\times$ downsampling sizes.

\subsection{Sparse Fusion Layer}
\label{sec3:sfl}

Lateral connections are often used to merge different levels of semantics~\cite{Lin2017FeaturePN} or dual-branch network~\cite{ChristophFeichtenhofer2018SlowFastNF}. Since the sparse voxel and pillar features are extracted separately via the 3D and 2D sparse convolutions, we introduce the Sparse Fusion Layer (SFL) to establish a bidirectional lateral connection for the element-wise fusion between the sparse voxel and pillar features. 

\subsubsection{Sparse Pooling and Broadcasting.} We apply the sparse pooling and broadcasting operations to match the size and dimension of sparse features before the fusion. Denote the sparse voxel features as voxel indices $\{v_i=(l_i, w_i, h_i) \in \mathbb{N}^{3} \}^{N_v}_{i=1}$ with corresponding features $\{ f^{v}_{i} \in \mathbb{R}^{D_v} \}^{N_v}_{i=1}$, where $N_v$ and $D_v$ refer to the number of non-empty voxels and voxel feature dimension. Likewise, sparse pillar indices and features are formed as $\{p_j=(l_j, w_j) \in \mathbb{N}^{2} \}^{N_p}_{j=1}$ and $\{ f^{p}_{j} \in \mathbb{R}^{D_p} \}^{N_p}_{j=1}$ respectively. $N_v$ is larger than $N_p$ due to many-to-one voxel-to-pillar correspondence. 

With the consistent occupancy in BEV assured, for a non-empty pillar in certain X-Y coordinate, we find the corresponding voxels by matching the horizontal vector of voxel indices $(l_i, w_i)$ with pillar indices $p_j=(l_j, w_j)$, and forms the $N_v\times N_p$ voxel-pillar index matrix $C$, where each element $c_{ij},i=1,\dots, N_v,j=1,\dots,N_p$ is defined as,

\begin{align}
  &c_{i j} =    
     \begin{cases}
          1, & l_i=l_j, w_i=w_j\\
          0, &\mathrm{otherwise}
     \end{cases} .
 \label{eq:indexing1}
\end{align}

Hence, the voxel-to-pillar sparse feature pooling and its inverse operation, feature broadcasting, are defined as,
\begin{equation}
  \text{Pool: }f^{v\rightarrow p}_{j} = \mathrm{Pool}\{f^{v}_{i}|c_{ij}=1, \forall i\},
  \label{eq:SparsePool}
\end{equation}

\begin{equation}
  \text{BroadCast: }f^{p\rightarrow v}_{i} = \{f^{p}_{j}| c_{ij}=1, \forall j \},
  \label{eq:BroadCast}
\end{equation}
where $\mathrm{Pool}\{\cdot\}$ denotes element-wise max-pooling and pillar features $f^{p}_{j}$ are identically projected onto the non-empty voxels in broadcasting.

\subsubsection{Sparse Voxel-Pillar Fusion.} In addition to sparse pooling and broadcasting, we further introduce the Sparse Fusion Layer (SFL), as illustrated in Fig.~\ref{fig:vpfusion}. Given the pair of sparse voxel and pillar features, SFL first calculates the voxel-pillar index matrix $C$. Then, sparse features engage in bidirectional interactions. In the voxel-to-pillar connection, voxel features undergo vertical aggregation by sparse max-pooling, detailed in Equ.~\ref{eq:SparsePool}, resulting in pillar-wise features with the shape $(N_p, D_v)$. These pillar-wise features are spatially consistent with original pillar features, while distinct in feature dimension. Next, a 2D submanifold convolution is applied to produce the pillar-wise pooled features $\{ f^{v\rightarrow p}_{j} \in \mathbb{R}^{D_p} \}^{N_p}_{j=1}$ with dimensions $(N_p, D_p)$. For pillar-to-voxel branch, pillar features are first transformed via the 2D submanifold convolution, and then broadcasted to form the voxel-wise features $\{ f^{p\rightarrow v}_{i} \in \mathbb{R}^{D_v} \}^{N_v}_{i=1}$ with the shape $(N_v, D_v)$, in Equ.~\ref{eq:BroadCast}. Notably, we adopt this asymmetrical structure with an emphasis on efficiency. Given that pillar's feature dimension is typically several times larger than that of the voxel, it's not cost-effective to directly transfer voxel features to the pillar dimension using the 3D submanifold convolution.  We then carry out element-wise summations for feature aggregation, 
\begin{equation}
  \begin{aligned}
  &f^{v}_{i} = f^{v}_{i} + f^{p\rightarrow v}_{i}, i=1,...,N_v, \\
  &f^{p}_{j} = f^{p}_{j} + f^{v\rightarrow p}_{j}, j=1,...,N_p.
  \end{aligned}
  \label{eq:spf_fusion}
\end{equation}

The SFL is incorporated after each intermediate step of the sparse Voxel-Pillar encoder to foster multi-level interactions between voxels and pillars. We highlight two primary benefits of the SFL. First, SFL offers a more straightforward approach for feature alignment by preserving the horizontal consistency in the sparse voxel-pillar encoder. Second, due to its dynamic nature, it avoids the random drops or paddings of voxels and pillars, thereby facilitating a transition to the fully sparse detection framework.

\subsection{Detection Framework}
\label{sec3:det_model}

Our method serves as a fully sparse backbone that can be seamlessly incorporated into different types of detectors. We introduce VPF$_\mathrm{de}$ and VPF$_\mathrm{sp}$, as variants of dense and sparse detectors respectively.

\subsubsection{VPF$_\mathrm{de}$.} For the dense grid-based methods, sparse voxel or pillar features are first converted to dense BEV feature maps and then processed by detection head. In this paper, we construct a dense detector named as VPF$_\mathrm{de}$. Given the sparse features from the proposed backbone, we first design the Dense Fusion Neck (DFN) to combine the dense feature maps from both voxel and pillar branches. DFN follows the common hierarchical structure~\cite{Lin2017FeaturePN, YanYan2018SECONDSE} for multi-scale feature aggregation. As shown in Fig.~\ref{fig:vpfusion}, we apply convolution blocks \texttt{Block}$(M, D)$ to extract dense features with $8 \times$ and $16 \times$ downsampling sizes in voxel/pillar branch separately, where $M$ and $D$ denote the number of convolution layers and output dimension. Next, dense voxels and pillar features with the same scale are fused by element-wise summation. Finally, we combine the different scale features via upsampling and concatenation as~\cite{YanYan2018SECONDSE}. For the detection head, we adopt the usual center-based head~\cite{yin2021center} with IoU-Aware rectification~\cite{hu2022afdetv2} to incorporate the regression accuracy with the classification score. The final predicted score is calculated by the rectification function~\cite{hu2022afdetv2},
\begin{equation}
  S_\mathrm{pred} = S^{1-\alpha}_\mathrm{cls} * \mathrm{IoU}^{\alpha}_\mathrm{pred},
  \label{eq:iou_rect}
\end{equation}
where $S_\mathrm{cls}$ is the classification score and $\mathrm{IoU}_\mathrm{pred}$ is the IoU prediction, $\alpha$ is the hyperparameter to balance the two. 

\subsubsection{VPF$_\mathrm{sp}$.} Sparse detection frameworks are presented to avoid redundant computation~\cite{chen2023voxelnext,fan2022embracing} and support long-range detection~\cite{LueFan2022FullyS3}. Since the proposed voxel-pillar encoder and SFL are fully sparse architectures, we could easily incorporate our hybrid representation backbone into the existing sparse detector, and therefore we present VPF$_\mathrm{sp}$. As demonstrated in Fig.~\ref{fig:vpfusion}, we apply additional down-sampling layers~\cite{chen2023voxelnext} for voxel and pillar branches, which obtain the $16 \times$ and $32 \times$ downsampled sparse features. These multi-scale sparse features are aggregated via the sparse height compression in each separate branch, and then combined with element-wise summation. Ultimately, the sparse head predicts objects from the sparse pillar-wise features in BEV space, and the box regression and IoU prediction paradigm stay the same as the center-based head and Equ.~\ref{eq:iou_rect}.

\subsubsection{Training Loss.}
\label{sec3:training_loss}

In our loss function, we use focal loss~\cite{lin2017focal} and L1 loss as the classification and box regression loss, noted $\mathcal{L}_\mathrm{cls}$ and $\mathcal{L}_\mathrm{reg}$ respectively. The IoU head is supervised by L1 loss and encoded by $(2*\mathrm{IoU}-0.5) \in [-1,1] $. We also use the Distance-IoU (DIoU) loss~\cite{zheng2020distance} to further optimize the object center regression,

\begin{equation}
  \mathcal{L}_\mathrm{diou} = 1-\mathrm{IoU}(b,b^{gt})+ \frac{c^2(b,b^{gt})}{d^2},
  \label{eq:iou_loss}
\end{equation}
where $\mathrm{IoU}(b,b^{gt})$ refers to the 3D IoU between predicted box $b$ and corresponding ground truth $b^{gt}$, $c$ denotes the center offset of $b$ and $b^{gt}$, and $d$ refers to the diagonal distance of minimum enclosing cuboid covering both $b$ and $b^{gt}$. 

Combined with the IoU prediction loss $\mathcal{L}_\mathrm{iou}$ and DIoU Loss $\mathcal{L}_\mathrm{diou}$, the overall loss is formed as,
\begin{equation}
  \mathcal{L} = \mathcal{L}_\mathrm{cls} + \mathcal{L}_\mathrm{iou} + \gamma( \mathcal{L}_\mathrm{diou} + \mathcal{L}_\mathrm{reg}),
  \label{eq:overall_loss}
\end{equation}
where $\gamma$ is the loss weight similar to~\cite{shi2022pillarnet}.

\section{Experiments}
\label{sec:experi}

\begin{table*}[!ht]
  \centering
    \resizebox{\textwidth}{!}{
    \begin{tabular}{c|c||c| c c c c | c c c c | c c c c}
      \hline
      \multirow{2}{*}{Methods} & \multirow{2}{*}{Stages}& LEVEL 2 & \multicolumn{4}{c|}{Veh. (LEVEL1/LEVEL2)} & \multicolumn{4}{c|}{Ped. (LEVEL1/LEVEL2)} & \multicolumn{4}{c}{Cyc. (LEVEL1/LEVEL2)}\\
      &  & mAP/mAPH & AP & APH & AP & \underline{APH} & AP & APH & AP & \underline{APH} & AP & APH & AP & \underline{APH} \\
      \hline
      SECOND~\shortcite{YanYan2018SECONDSE}& Single& 61.0/57.2 &72.3& 71.7& 63.9& 63.3& 68.7& 58.2& 60.7& 51.3 & 60.6 & 59.3 & 58.3 & 57.0 \\
      MVF~\shortcite{zhou2020end} & Single& - & 62.9 & - & - & - & 65.3 & -  & - & - & - & - & - & -  \\
      PointPillars~\shortcite{lang2019pointpillars} & Single& 62.8/57.8 & 72.1 & 71.5 & 63.6 & 63.1 & 70.6 & 56.7 & 62.8 & 50.3 & 64.4 & 62.3 & 61.9 & 59.9 \\
      Pillar-OD~\shortcite{wang2020pillar} & Single& - & 69.8 & - & -  & - & 72.5 & -  & - & - & - & - & - & - \\
      AFDetV2-Lite~\shortcite{hu2022afdetv2} & Single& 71.0/68.8 & 77.6 & 77.1 & 69.7 & 69.2 & 80.2 & 74.6 & 72.2 & 67.0 & 73.7 & 72.7 & 71.0 & 70.1 \\
      IA-SSD~\shortcite{YifanZhang2022NotAP} & Single& 62.3/58.1 & 70.5 & 69.7 & 61.6 & 61.0 & 69.4 & 58.5 & 60.3 & 50.7 & 67.7 & 65.3 & 65.0 & 62.7 \\
      CenterFormer~\shortcite{zhou2022centerformer} & Single& 71.1/68.9 & 75.0 & 74.4 & 69.9 & 69.4 & 78.6 & 73.0 & 73.6 & 68.3 & 72.3 & 71.3 & 69.8 & 68.8\\
      Centerpoint$^{\dagger}$~\shortcite{yin2021center}& Single& 71.7/69.4 &77.7&77.1&69.7&69.2&80.9&75.1&72.6&67.2&75.6&74.4&72.8&71.7\\
      PillarNet~\shortcite{shi2022pillarnet} & Single& 71.0/68.5 & 79.1 & 78.6 & 70.9 & 70.5 & 80.6 & 74.0 & 72.3 & 66.2 & 72.3 & 71.2 & 69.7 & 68.7 \\
      VoxelNeXt2D~\shortcite{chen2023voxelnext} & Single&70.9/68.2& 77.9 & 77.5 & 69.7 & 69.2 & 80.2 & 73.5 & 72.2 & 65.9 & 73.3 & 72.2 & 70.7 & 69.6\\ 
      VoxelNeXt$_{K3}$~\shortcite{chen2023voxelnext} &Single&72.2/70.1 &78.2&77.7 &69.9&69.4 &81.5&76.3 &73.5&68.6 &76.1&74.9 &73.3&72.2\\
      GD-MAE~\shortcite{yang2023gd} & Single & 72.9/70.4 & 79.4 & 78.9 & 70.9 & 70.5 & 82.2 & 75.9 & 74.8 & 68.8 & 75.8 & 74.8 & 73.0 & 72.0 \\
      DSVT~\shortcite{wang2023dsvt} & Single& 73.2/71.0 & 79.3 & 78.8 & 70.9 & 70.5 & 82.8 & 77.0 & \textbf{75.2} & 69.8 & 76.4 & 75.4 & 73.6 & 72.7 \\
      \hline
      VPF$_\mathrm{sp}$-Lite&\multirow{4}{*}{Single} & 73.1/70.9 & 79.1 & 78.6 & 71.1 & 70.6 & 82.1 & 76.7 & 74.5 & 69.4 & 76.5 & 75.5 & 73.6 & 72.7\\
      VPF$_\mathrm{de}$-Lite & & 73.6/71.4 & 80.2 & 79.7 & 71.9 & 71.5 & 82.5 & 76.9 & 74.8 & 69.4 & 77.1 & 76.0 & 74.2 & 73.2 \\
      VPF$_\mathrm{sp}$& & \textbf{73.6/71.6} & \textbf{80.2} & \textbf{79.8} & \textbf{71.7} & \textbf{71.3} & \textbf{82.5} & \textbf{77.5} & 74.9 & \textbf{70.1} & \textbf{77.2} & \textbf{76.2} & \textbf{74.3} & \textbf{73.3}\\
      VPF$_\mathrm{de}$ & & \textbf{73.9/71.7} & \textbf{80.5} & \textbf{80.0} & \textbf{72.3} & \textbf{71.9} & \textbf{82.8} & \textbf{77.3} & \textbf{75.1} & \textbf{69.9} & \textbf{77.2} & \textbf{76.0} & \textbf{74.3} & \textbf{73.2} \\
      \hline
      Voxel RCNN~\shortcite{JiajunDeng2021VoxelRT} & Two& - & 75.6 & - & 66.6 & - & - & - & - & - & - & - & - & - \\
      PartA$^2$~\shortcite{shi2020points} & Two& 66.9/63.8 & 77.1 & 76.5 & 68.5 & 68.0 & 75.2 & 66.9  & 66.2 & 58.6 & 68.6 & 67.4 & 66.1 & 64.9 \\
      LiDAR R-CNN~\shortcite{li2021lidar} & Two& 65.8/61.3 & 76.0 & 75.5 & 68.3 & 67.9 & 71.2 & 58.7 & 63.1 & 51.7 & 68.6 & 66.9 & 66.1 & 64.4 \\
      RSN~\shortcite{sun2021rsn} & Two& - & 75.1 & 74.6 & 66.0 & 65.5 & 77.8 & 72.7 & 68.3 & 63.7 & - & - & - & - \\
      PV-RCNN~\shortcite{shi2020pv} & Two& 66.8/63.3 & 77.5 & 76.9 & 69.0 & 68.4 & 75.0 & 65.6 & 66.0 & 57.6 & 67.8 & 66.4 & 65.4 & 64.0 \\
      SST\_TS\_1f~\shortcite{fan2022embracing} & Two& - & 76.2 & 75.8 & 68.0 & 67.6 & 81.4 & 74.0 & 72.8 & 65.9 & - & - & - & - \\
      PV-RCNN++~\shortcite{shi2022pv} & Two& 71.7/69.5 & 79.3 & 78.8 & 70.6 & 70.2 & 81.3 & 76.3 & 73.2 & 68.0 & 73.7 & 72.7 & 71.2 & 70.2 \\
      FSD$_{spconv}$~\shortcite{LueFan2022FullyS3} & Two& 72.9/70.8 & 79.2 & 78.8 & 70.5 & 70.1 & 82.6 & 77.3 & 73.9 & 69.1 & 77.1 & 76.0 & 74.4 & 73.3 \\
      GD-MAE~\shortcite{yang2023gd}&Two&74.1/71.6& 80.2&79.8 &72.4&72.0& 83.1&76.7& 75.5&69.4& 77.2&76.2& 74.4&73.4\\
      DSVT~\shortcite{wang2023dsvt} & Two&74.3/72.1&80.2&79.7&72.0&71.6&83.7&78.0&76.1&70.7&\textbf{77.8}&\textbf{76.8}&\textbf{74.9}&\textbf{73.9}\\
      \hline
      VPF$_\mathrm{sp}$-TS& \multirow{2}{*}{Two} & \textbf{74.7/72.5} & \textbf{81.4} & \textbf{80.9} & \textbf{73.1} & \textbf{72.7} & \textbf{83.7} & \textbf{78.5} & \textbf{76.2} & \textbf{71.3} & 77.3 & 76.3 & 74.4 & 73.5\\
      VPF$_\mathrm{de}$-TS & & \textbf{75.1/72.9} & \textbf{81.4} & \textbf{80.9} & \textbf{73.6} & \textbf{73.2} & \textbf{84.0} & \textbf{78.6} & \textbf{76.6} & \textbf{71.3} & \textbf{77.8} & \textbf{76.7} & \textbf{75.2} & \textbf{74.1}\\
      \hline
    \end{tabular}}
    \caption{Single-frame performance comparison on the WOD {\em val} set, without test-time augmentation or model ensemble. $\;^{\dagger}$: single-stage with IoU prediction setting. Top-2 results are highlighted in bold for different stages. } 
  \label{tab:waymo-val}
\end{table*}

\subsection{Datasets}

\noindent{\textbf{Waymo Open Dataset}} (WOD)~\cite{sun2020scalability} consists of 798 training, 202 validation, and 150 testing sequences with 200K annotated frames. The detection range covers the area of $[-75m,-75m, 75m,75m]$, and evaluation metrics are average precision (AP) and average precision weighted by heading (APH). Test samples are split into two difficulties, LEVEL 1 for objects with more than 5 inside points and LEVEL 2 for objects with at least 1 point.

\noindent{\textbf{nuScenes Dataset}}~\cite{caesar2020nuscenes} contains 700 training, 150 validation, and 150 testing sequences, with 34K annotated keyframes and 10 categories. The evaluation metric is nuScenes detection score (NDS), which is a weighted sum of mAP and other true positive metrics.

\subsection{Implementation Details}

\subsubsection{Network Architecture.}

Our VPF builds on the sparse backbone, which consists of the sparse voxel-pillar encoder with $4$ convolution blocks and Sparse Fusion Layers (SFL). In the VPF$_\mathrm{de}$ model, the feature dimensions for the voxel and pillar branches in each convolution block are set at $[16,32,64,64]$ and $[32,64,128,256]$, respectively. Meanwhile, for the VPF$_\mathrm{sp}$ model, these dimensions are $[16,32,64,128]$ and $[32,64,128,256]$. 

Additionally, for the Dense Fusion Neck~(DFN) applied in our dense detectors, the number of convolution layers $M$ and the output dimension $D$ are set as $5$, $128$. As for our sparse detectors, feature dimensions in additional downsampled layers ({\em i.e.}, 16$\times$ and 32$\times$ convolution blocks) are set as $[128,128]$, $[256, 256]$ for the voxel and pillar branches. Note that the voxel features are upsampled to $256D$ during the final voxel-pillar feature fusion.

\subsubsection{Model Setting.} In data preprocessing, point clouds are divided into voxels and pillars with the size of $[0.1m, 0.1m, 0.15m]$, $[0.1m, 0.1m]$ for WOD, and $[0.075m, 0.075m, 0.2m]$, $[0.075m, 0.075m]$ for nuScenes dataset. We apply the IoU thresholds $(0.8, 0.55, 0.55)$ and $\alpha$ in Equ.~\ref{eq:iou_rect} as $(0.68, 0.71, 0.65)$ for Vehicle, Pedestrian, and Cyclist in WOD. In nuScenes, the IoU and $\alpha$ are set as $0.2$ and $0.5$ for all categories.

\subsubsection{Training Details.} All models are trained on 4 NVIDIA 3090 GPUs with Adam optimizer. The learning rates are set as 1e-3 and 3e-3 for nuScenes and Waymo dataset, separately. We use the common data augmentation strategies~\cite{shi2022pv} and ground-truth sampling fade strategy as~\cite{wang2023dsvt}. For WOD, we train VPF$_\mathrm{de}$ and VPF$_\mathrm{sp}$ with batch size 16 for 15 epochs and 12 epochs separately. For nuScenes, our models are trained with the same batch size for 20 epochs.

\subsection{Comparison with the State-of-the-art}

\subsubsection{Waymo Open Dataset.} 
\begin{figure}[tbp]
	\centering
	\includegraphics[width=\linewidth]{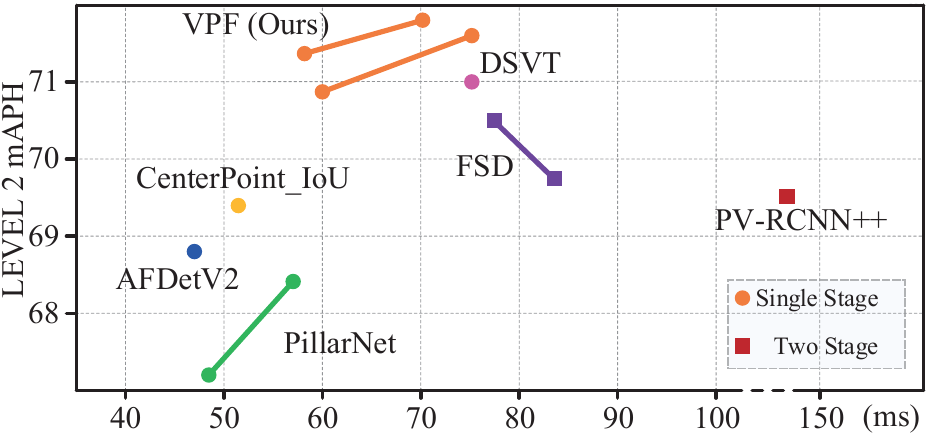}
  \caption{Performance vs. inference latency on WOD {\em val} set. Tested on a single 3090 GPU with batch size $1$.}
	\label{fig:infer_speed}
  \vspace{-0.15in}
\end{figure}

We compare our VPF variants with existing single-frame methods on WOD {\em val} set. As shown in Tab.~\ref{tab:waymo-val}, for the single-stage framework, our method achieves promising performance among all single-stage methods. Specifically, both VPF$_\mathrm{sp}$ and VPF$_\mathrm{de}$ are superior to dense or sparse baselines (i.e., Centerpoint, PillarNet, and VoxelNeXts) with remarkable margins. Moreover, our method outperforms recent transformer-based GD-MAE~\cite{yang2023gd} and DSVT~\cite{wang2023dsvt} with more efficient training schedule (4 3090 vs. 8 A100 GPUs or MAE pre-training) and inference speed. We also provide lightweight versions of our framework, VPF$_\mathrm{sp}$-Lite and VPF$_\mathrm{de}$-Lite, which reduce the convolution layer and channel dimension for a lower computing budget. They achieve competitive results while running at 60 ($\pm$5) ms and 58 ($\pm$4) ms, as demonstrated in Fig.~\ref{fig:infer_speed}. To compare with two-stage detectors, we simply adopt CT3D~\cite{Sheng2021Improving3O} as the second stage, obtaining competitive performance as well. While the performance of VPF$_\mathrm{sp}$ is inferior to VPF$_\mathrm{de}$ in the Vehicle category since it's harder to estimate the object center estimation by sparse architecture, VPF$_\mathrm{sp}$ is more computationally efficient in long-range detection.

\subsubsection{nuScenes Dataset.} We evaluate and compare VPF with LiDAR-based methods on the nuScenes {\em test} set. As depicted in Tab.~\ref{tab:nus_test}, our approach surpasses previous methods in both mAP and NDS metrics with a considerable improvement. More specifically, VPF$_\mathrm{de}$ achieves $67.0$ mAP and $72.7$ NDS in the single model setting, with the NDS boosts to \textbf{$73.8$} using double-flip test. In addition, VPF improves the performance of vertical sensitive categories (e.g., Pedestrian, Bicycle) by a large margin while maintaining competitive results on other classes (e.g., Truck). We attribute this to our robust vertical representation. Given that slender objects, like pedestrians, present richer vertical information than horizontal, our hybrid voxel-pillar encoding paradigm is helpful in establishing effective representations for such entities.

\begin{table*}[t!]
  \centering
  \tiny
  \resizebox{\textwidth}{!}{
  \begin{tabular}{c|c||c|cccccccccc}
  \hline
  {Method} &  NDS & mAP  & Car & Truck & Bus & Trailer & C.V. & Ped & Mot & Byc & T.C. & Bar \\ \hline 
  PointPillars~\shortcite{lang2019pointpillars} & 45.3 & 30.5 & 68.4 & 23.0 & 28.2 & 23.4 & 4.1 & 59.7 & 27.4 & 1.1 & 30.8 & 38.9 \\ 
  3DSSD~\shortcite{ZetongYang20203DSSDP3} &  56.4 & 42.6 &  81.2 & 47.2 & 61.4 & 30.5 & 12.6 & 70.2 & 36.0 & 8.6 & 31.1 & 47.9 \\ 
  CBGS~\shortcite{zhu2019class} &  63.3 & 52.8 & 81.1 & 48.5 & 54.9 & 42.9 & 10.5 & 80.1 & 51.5 & 22.3 & 70.9 & 65.7 \\ 
  HotSpotNet~\shortcite{chen2020object} &  66.0 & 59.3 & 83.1 & 50.9 & 56.4 & 53.3 & 23.0 & 81.3 & 63.5 & 36.6 & 73.0 & 71.6 \\
  CVCNET~\shortcite{chen2020every}       &  66.6  & 58.2 & 82.6 & 49.5 & 59.4 & 51.1 & 16.2 & 83.0 & 61.8 & 38.8 & 69.7 & 69.7 \\ 
  CenterPoint~\shortcite{yin2021center}  &  65.5 & 58.0 & 84.6 & 51.0 & 60.2 & 53.2 & 17.5 & 83.4 & 53.7 & 28.7 & 76.7 & 70.9 \\ 
  CenterPoint$^{\dagger}$~\shortcite{yin2021center} &  67.3 & 60.3 & 85.2 & 53.5 & 63.6 & 56.0 & 20.0 & 84.6 & 59.5 & 30.7 & 78.4 & 71.1 \\ 
  AFDetV2-Lite~\shortcite{hu2022afdetv2} & 68.5 & 62.4 & 86.3 & 54.2 & 62.5 & 58.9 & 26.7 & 85.8 & 63.8 & 34.3 & 80.1 & 71.0 \\
  VISTA-OHS$^{\dagger}$~\shortcite{deng2022vista} & 69.8 & 63.0 & 84.4 & 55.1 & 63.7 & 54.2 & 25.1 & 84.6 & 70.0 & 45.4 & 78.5 & 71.1 \\
  Focals Conv~\shortcite{chen2022focal}  &   70.0 & 63.8 & \underline{86.7} & \underline{56.3} & \underline{\textbf{67.7}} & 59.5 & 23.8 & 87.5 & 64.5 & 36.3 & 81.4 & 74.1 \\ 
  LargeKernel3D~\shortcite{chen2022scaling} &   70.5 & 65.3 & 85.9 & 55.3 & 66.2 & 60.2 & 26.8 & 85.6 & 72.5 & 46.6 & 80.0 & 74.3 \\
  PillarNet-34$^{\dagger}$~\shortcite{shi2022pillarnet}  &   71.4 & 66.0 & \textbf{87.6} & 57.5 & 63.6 & 63.1 & 27.9 & 87.3 & 70.1 & 42.3 & 83.3 & 77.2 \\ 
  VoxelNeXt$^{\dagger}$~\shortcite{chen2023voxelnext} & 71.4& 66.2 &85.3 &55.7 &66.2 &57.2 &29.8& 86.5& 75.2& 48.8& 80.7& 76.1\\
  LinK~\cite{lu2023link} & 71.0& 66.3 & 86.1 &55.7& 65.7 &62.1& 30.9 &85.8& 73.5 &47.5& 80.4 & 75.5\\
  \hline
  VPF$_\mathrm{de}$ &  \underline{72.7} & \underline{67.0} & 85.8 & 55.1 & 63.5 & \underline{62.1} & \underline{33.3} & \underline{87.6} & \underline{72.5} & \underline{48.6} & \underline{82.9} & \textbf{\underline{78.2}} \\ 
  VPF$_\mathrm{de}$$^{\dagger}$ &  \textbf{73.8} & \textbf{68.6} & 86.3 & \textbf{56.8} & 66.1 & \textbf{64.5} & \textbf{34.6} & \textbf{88.3} & \textbf{75.8} & \textbf{51.9} & \textbf{84.2} & 77.6 \\
  \hline
  \end{tabular}}
  \caption{Performance comparison for 3D object detection on the nuScenes {\em test} set. $\;^{\dagger}$ indicates the flipping test is used. We highlight the best results in bold and the best non-ensemble results with the underline.}
  \label{tab:nus_test}
\end{table*}

\begin{table}[t!]
  \centering
  \Large
  \resizebox{\linewidth}{!}{
  \begin{tabular}{ c c c c|c|c c c c}
  \hline
   \multirow{2}{*}{Voxel} & \multirow{2}{*}{Pillar} & \multirow{2}{*}{DFN} & \multirow{2}{*}{SFL}& \multirow{2}{*}{Lat.} & \multicolumn{4}{c}{LEVEL 2 APH} \\
   & & & && Veh. & Ped. & Cyc. & mAPH \\
  \hline
   \checkmark &  &  &  & 51ms & 67.8 & 66.5 & 68.2 & 67.5 \\
    & \checkmark &  &  & 57ms &68.5 & 64.1 & 66.2 & 66.3 \\
   \checkmark & \checkmark &  &  & 102ms &69.1 & 66.7 & 69.2 & 68.3 \\
   \checkmark & \checkmark & \checkmark &  & 69ms &69.2 & 66.0 & 68.9 & 68.0 \\ 
   \checkmark & \checkmark & \checkmark & \checkmark & 75ms &\textbf{70.2} & \textbf{68.2} & \textbf{70.7} & \textbf{69.7} \\  
  \hline
  \end{tabular}}
  \caption{Effects of hybrid encoding paradigm. CenterPoint and PillarNet are used as voxel/pillar-based baselines.}
  \label{tab:component}
\end{table}

\begin{table}[t!]
  \centering
  \small
  \resizebox{\linewidth}{!}{
  \begin{tabular}{c|c|c c c c}
  \hline
  \multirow{2}{*}{Exp.} & \multirow{2}{*}{Param.} & \multicolumn{4}{c}{LEVEL 2 APH} \\
  & & Veh. & Ped. & Cyc. & mAPH \\
  \hline
  Scaled PillarNet & 92.4M & 69.0 & 64.8 & 67.8 & 67.2 \\
  Scaled CenterPoint & 93.2M & 69.2 & 67.0 & 70.1 & 68.8\\
  Scaled CenterPoint$^{\dagger}$ & 96.8M & 69.2 & 67.2 & 70.5 & 69.0\\
  Voxel-Voxel & 67.1M & 68.9 & 66.2 & 69.5 & 68.2\\
  Pillar-Pillar & 111.1M & 69.3 & 64.5 & 67.1 & 66.9\\
  VPF$_\mathrm{de}$ & 92.4M & \textbf{70.2} & \textbf{68.2} & \textbf{70.7} & \textbf{69.7} \\  
  \hline
  \end{tabular}}
  \caption{Effects of architecture and scalability. $\;^{\dagger}$: deploy the vertical residual for feature aggregation.}
  \label{tab:effect_single}
\end{table}

\subsection{Ablation Studies}
\label{sec4.4:ablation}
In this section, we conduct experiments on WOD {\em val} set to validate the effectiveness of voxel-pillar hybrid representation. Moreover, we investigate the impact of varying point cloud representations and architectures on 3D detection. All models are trained on the full train set for 7 epochs, with comparisons based on APH of LEVEL 2 difficulty.

\subsubsection{Effect of Voxel-Pillar Encoding Paradigm.} We take VPF$_\mathrm{de}$ as the framework to illustrate the effects of hybrid representation. For voxel-/pillar-based baselines, we deploy the single-stage Centerpoint~\cite{yin2021center} and PillarNet~\cite{shi2022pillarnet}. To ensure a fair comparison, we add the IoU prediction to CenterPoint as~\cite{shi2022pillarnet}. First, we compare the single representations with the naive model ensemble, in which the predictions from CenterPoint and PillarNet are combined via NMS, as shown in the $3^{rd}$ row. The ensemble strategy consistently improves the detection accuracy. Then, as presented in the $4^{th}$ row, we deploy the proposed voxel-pillar encoder with Dense Fusion Neck (DFN). This late fusion strategy achieves comparable results while being more efficient due to the unified encoding process of voxels and pillars. Finally, we deploy the Sparse Fusion Layer (SFL), obtaining $1.7$ mAPH improvement with a minor increase in latency, which demonstrates the importance of voxel-pillar interaction within the sparse backbone.

\begin{table}[t!]
  \centering
  \scriptsize
  \resizebox{\linewidth}{!}{
  \begin{tabular}{c| l | c c c c}
  \hline
  \multirow{2}{*}{Exp.} & \multicolumn{1}{c|}{\multirow{2}{*}{Connections}} & \multicolumn{4}{c}{LEVEL 2 APH} \\
  & & Veh. & Ped. & Cyc. & mAPH \\
  \hline
  1 & Voxel-to-Pillar & 70.0 & 67.3 & 70.4 & 69.2 \\
  2 & Pillar-to-Voxel & 69.4 & 66.3 & 68.7 & 68.1 \\
  3 & Bidirectional & \textbf{70.2} & \textbf{68.2} & \textbf{70.7} & \textbf{69.7}\\
  \hline
  \end{tabular}}
  \caption{Ablation on different interaction paradigm. }
  \label{tab:sfl_connect}
\end{table}

\begin{figure}[tbp]
  \centering
  \includegraphics[width=\linewidth]{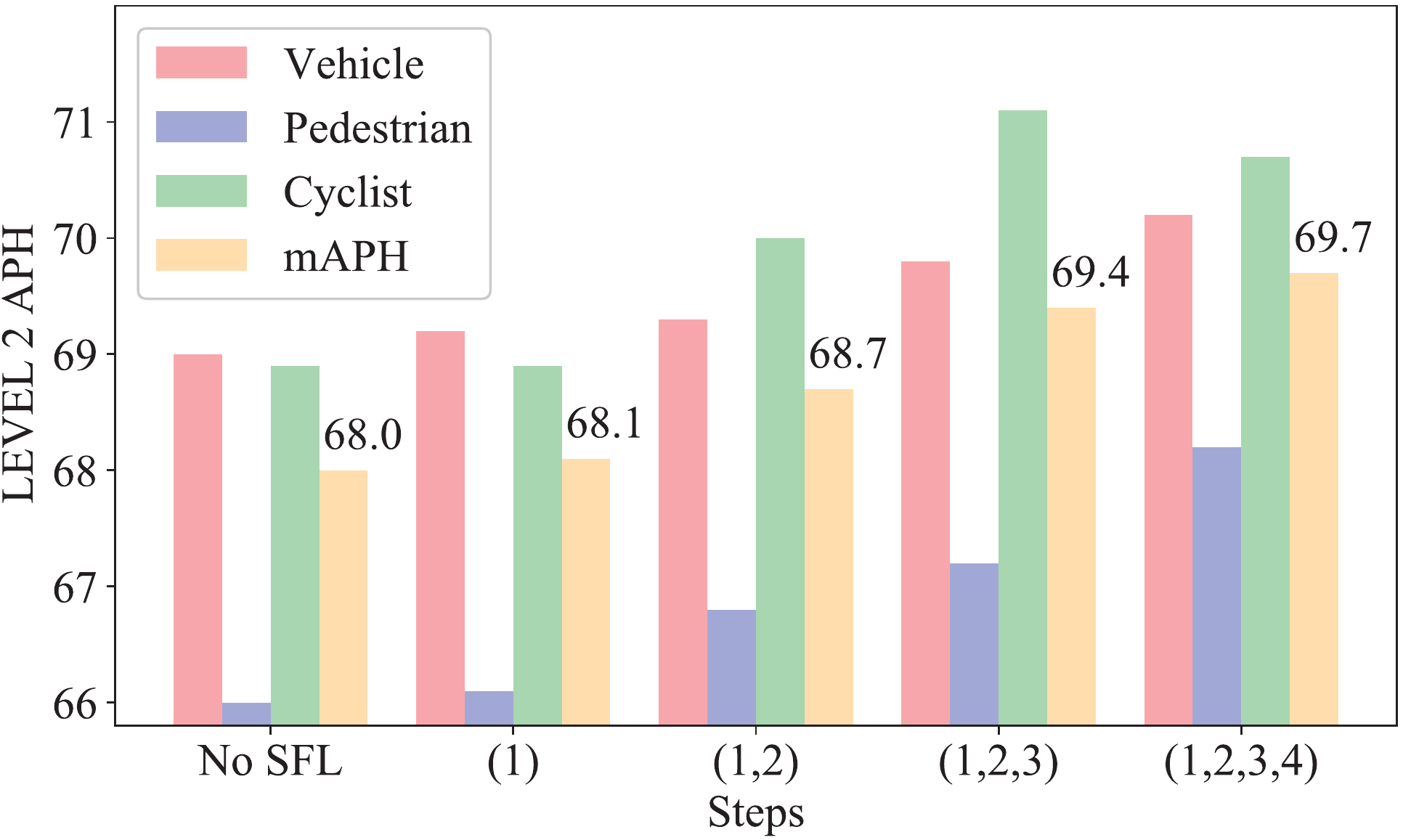}
  \caption{Ablation on deploying steps for SFL.}
  \label{fig:sfl_stages}
\end{figure}

\subsubsection{Effects of Model Scale and Architecture.} Since our model has a different scale and dual-branch pipeline, we ablate the effectiveness of model scale and architecture in Tab.~\ref{tab:effect_single}. First, we construct the scaled PillarNet and CenterPoint by compositely increasing the number of layers and feature dimensions while keeping the BEV map resolution unchanged, which makes them comparable with VPF in parameter scales. As shown in the $1^{st}-2^{nd}$ rows, our hybrid voxel-pillar encoding paradigm is significantly prior to the single representation counterpart. Then, we explore the validity of vertical feature aggregation in the $3^{rd}$ row. For each stage within the voxel backbone, we introduce a vertical residual. This residual is constructed by aggregating voxel features vertically via sparse pooling. The gathered features serve as residuals and are added to the original voxel features. This vertical residual leads to an improvement on slender objects, {\em i.e.}, Pedestrian and Cyclist, while it still has a large gap compared with our approach. In addition, we investigate different dual-branch pipelines to explore the impact from wider or multi-branch networks~\cite{chen2017dual, xie2017aggregated}. The results of voxel-only or pillar-only frameworks in $4^{th}-5^{th}$ rows of Tab.~\ref{tab:effect_single} indicates although multi-branch architecture could boost detection performance, the improvements in our work are mainly from the synergy of voxel and pillar.

\subsubsection{Sparse Fusion Layer.} We study the effects of the SFL from two perspectives. First, we verify the bidirectional connection of SFL. Given its voxel-to-pillar and pillar-to-voxel connections, we compare the performance of the interaction paradigm in Tab.~\ref{tab:sfl_connect}. It indicates that each individual connection could obtain clear improvement, and applying the bidirectional connection achieves the best results. Second, we analyze the deploying intermediate steps of SFL for the voxel-pillar encoder in Fig.~\ref{fig:sfl_stages}. The performance increases with the module stacking in each step. As SFL amplifies the local context for the pillar branch and infuses the vertical semantics into the voxel branch, it primarily functions in the latter steps of the sparse backbone, helping to enrich the vertical information during sparse convolutions.

\section{Conclusion}
\label{sec:conclusion}
Inspired by the distinctions and constraints of voxel and pillar encoding paradigms, we propose the VPF, a hybrid detection framework that combines the strengths of both. VPF utilizes a sparse voxel-pillar encoder for consistent dual-branch feature extraction. Further, our Sparse Fusion Layer facilitates bidirectional interaction between the sparse voxel and pillar features, jointly enhancing vertical representation learning. Notably, our proposed components can be seamlessly integrated into both dense and sparse detectors, yielding promising performance and real-time inference speeds.
\hspace*{\fill}

\noindent\textbf{Limitations.} VPF relies on consistent occupancy in the bird's eye view for element-wise fusion, requiring equal horizontal partitioning for voxels and pillars. For future work, we will explore the asynchronous voxel-pillar fusion strategies (e.g., varying resolution or multi-stride fusion) for more flexible point cloud representation learning.


\section*{Acknowledgements}
This paper is supported by the National Natural Science Foundation of China (No. 62088102). We would like to thank four anonymous reviewers for their constructive comments.

\bibliography{aaai24}
\clearpage
\appendix

\renewcommand\thefigure{\Alph{section}\arabic{figure}}    
\renewcommand\thetable{\Alph{section}\arabic{table}}

\begin{figure*}[ht!]
    \centering
    \small
    \newlength{\tempdima} 
    \setlength{\tempdima}{0.32\linewidth} 
    \begin{minipage}[t]{\tempdima}
        \centering
        \includegraphics[width=\linewidth,trim=10 10 10 10,clip]{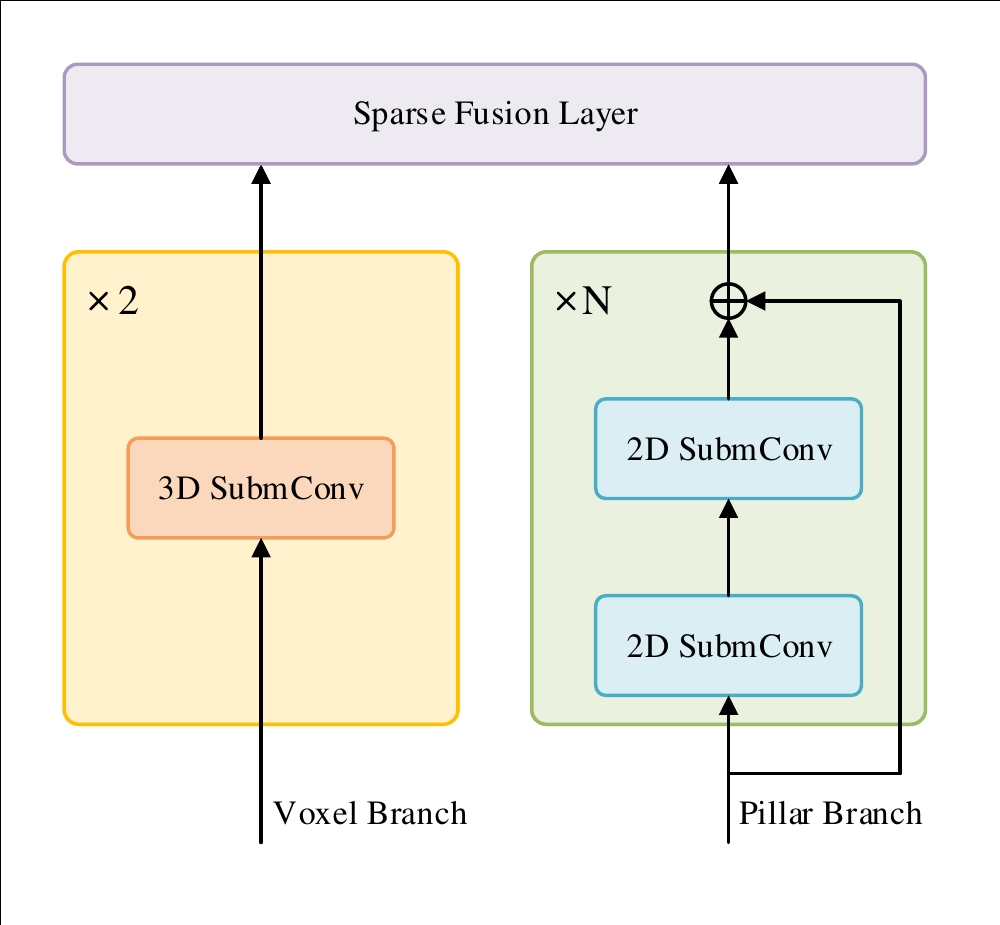}
        \footnotesize{(a) VPF$_\mathrm{de}$}
    \end{minipage}
    \hfill 
    \begin{minipage}[t]{\tempdima}
        \centering
        \includegraphics[width=\linewidth,trim=10 10 10 10,clip]{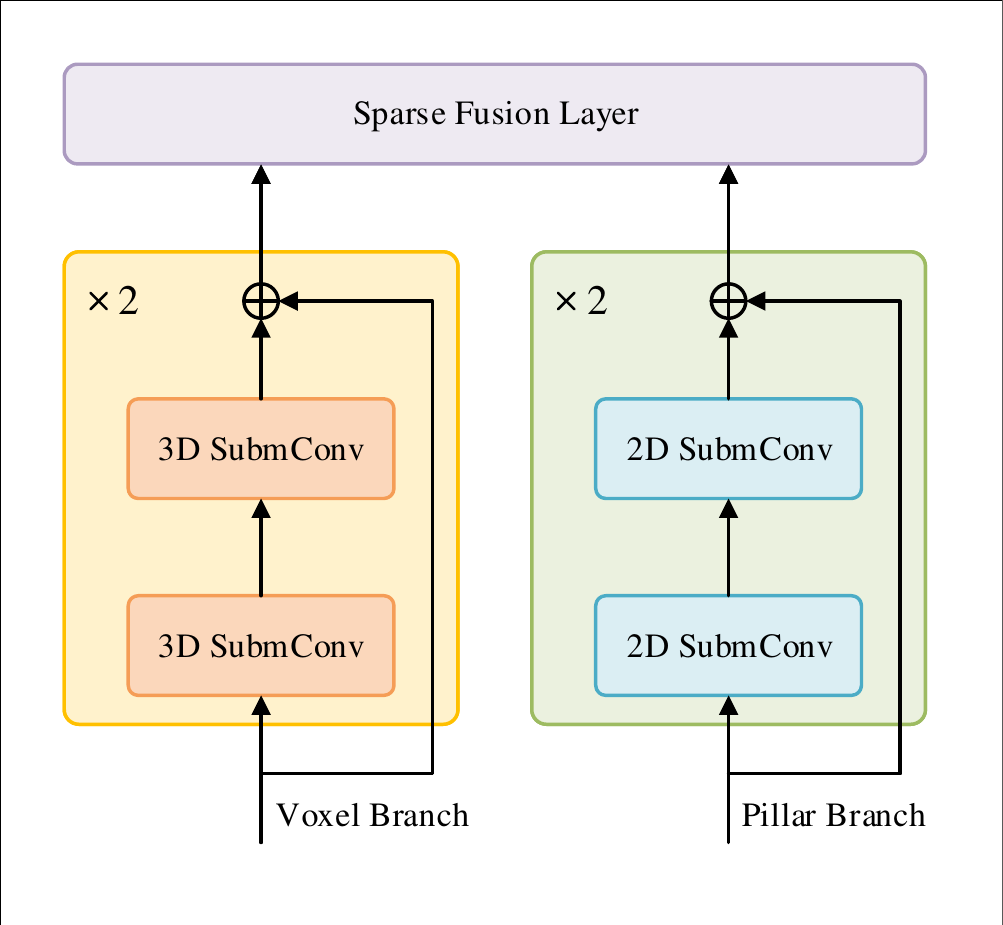}
        \footnotesize{(b) VPF$_\mathrm{sp}$}
    \end{minipage}
    \hfill 
    \begin{minipage}[t]{\tempdima}
        \centering
        \includegraphics[width=\linewidth,trim=10 10 10 10,clip]{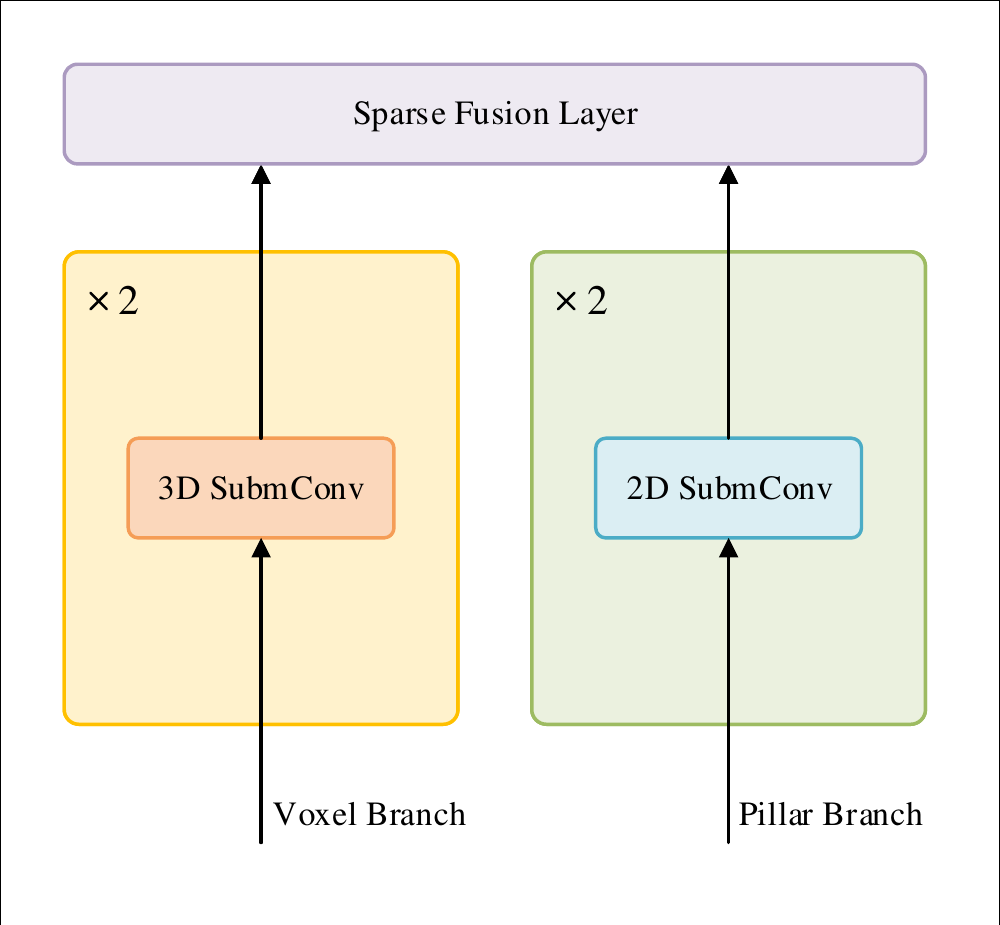}
        \footnotesize{(c) VPF Lite}
    \end{minipage}
    \caption{Model architectures for VPF variants. (a), (b) and (c) illustrate the dual-branch sparse conv blocks utilized in VPF$\mathrm{de}$, VPF$\mathrm{sp}$, all VPF lite models, respectively.}
    \label{fig:vpf_structure}
\end{figure*}

\section{More Implementation Details}

\subsection{Network Structure.}

Our VPF builds on the sparse backbone, which consists of the sparse voxel-pillar encoder with $4$ conv blocks and Sparse Fusion Layers (SFL). In the VPF$_\mathrm{de}$ model, the feature dimensions for the voxel and pillar branches in each convolution block are set at $[16,32,64,64]$ and $[32,64,128,256]$, respectively. Meanwhile, for the VPF$_\mathrm{sp}$ model, these dimensions are $[16,32,64,128]$ and $[32,64,128,256]$. The different convolution block structures for VPF variants are illustrated in Fig.~\ref{fig:vpf_structure}. 

Additionally, for the Dense Fusion Neck~(DFN) applied in our dense detectors, the number of convolution layers $M$ and the output dimension $D$ are set as $5$, $128$. As for our sparse detectors, feature dimensions in additional downsampled layers ({\em i.e.}, 16$\times$ and 32$\times$ conv blocks) are set as $[128,128]$, $[256, 256]$ for the voxel and pillar branches. Note that the voxel features are upsampled to $256D$ during the final voxel-pillar feature fusion.

\subsection{Training Settings.} 

We set the learning rate as 1e-3 and 3e-3 for nuScenes and Waymo dataset, separately. For the two-stage models, we follow DSVT~\cite{wang2023dsvt} to fix the first stage of the detection model and finetune CT3D~\cite{Sheng2021Improving3O} refinement stage for 12 epochs.

\section{Experimental Results}

\subsection{Vertical Density Analysis}

\begin{figure}[t!]
\centering
\small
\centerline{
    \begin{minipage}[t]{0.49\linewidth}
    \centerline{
        \includegraphics[width=\linewidth,trim=14 0 33 18,clip]{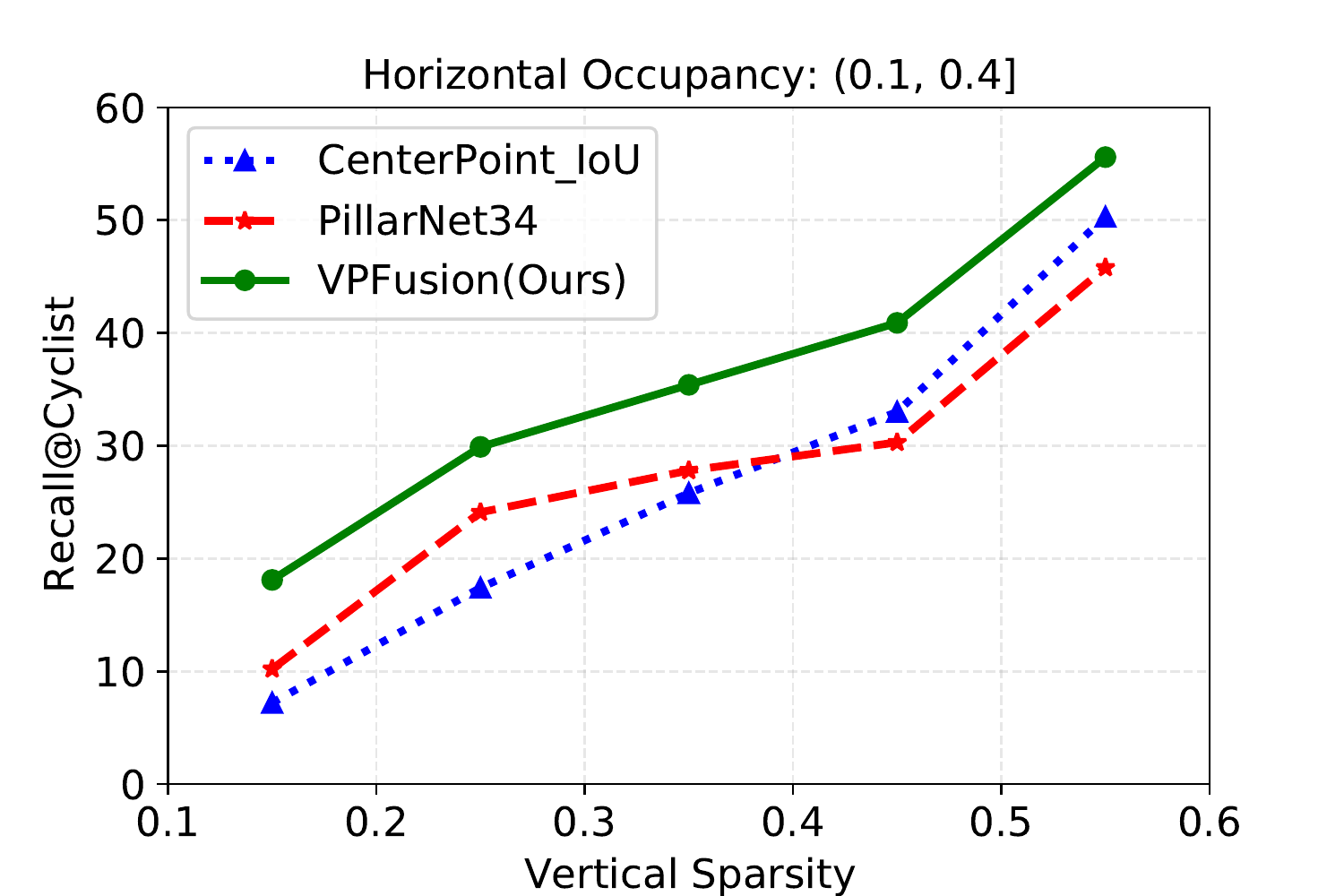}
    }
    \end{minipage}
    \begin{minipage}[t]{0.49\linewidth}
    \centerline{
        \includegraphics[width=\linewidth,trim=14 0 33 18,clip]{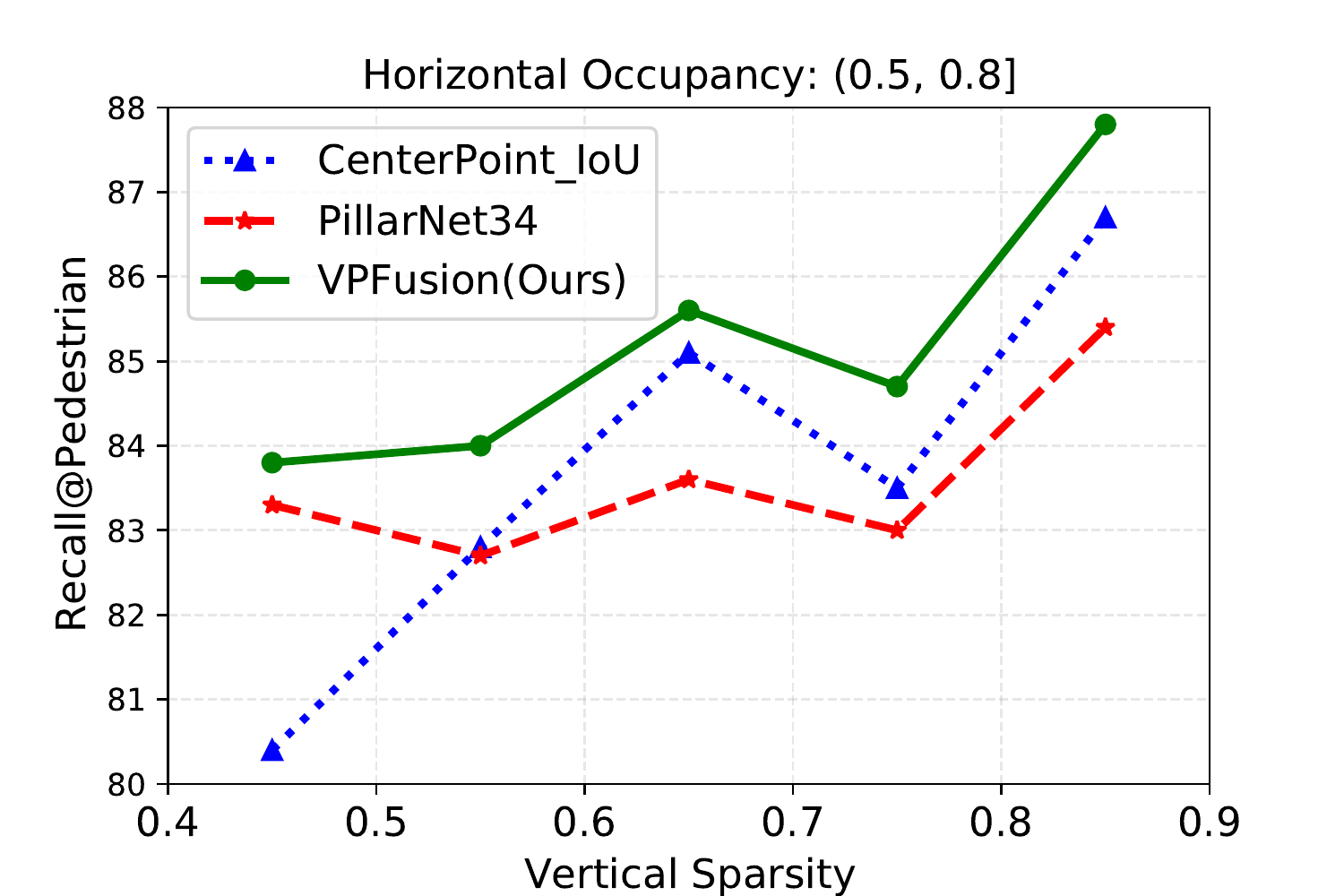}
    }
    \end{minipage}
}
\vspace{2pt}
\centerline{
    \begin{minipage}[t]{0.49\linewidth}
    \centerline{
        \includegraphics[width=\linewidth,trim=14 0 33 18,clip]{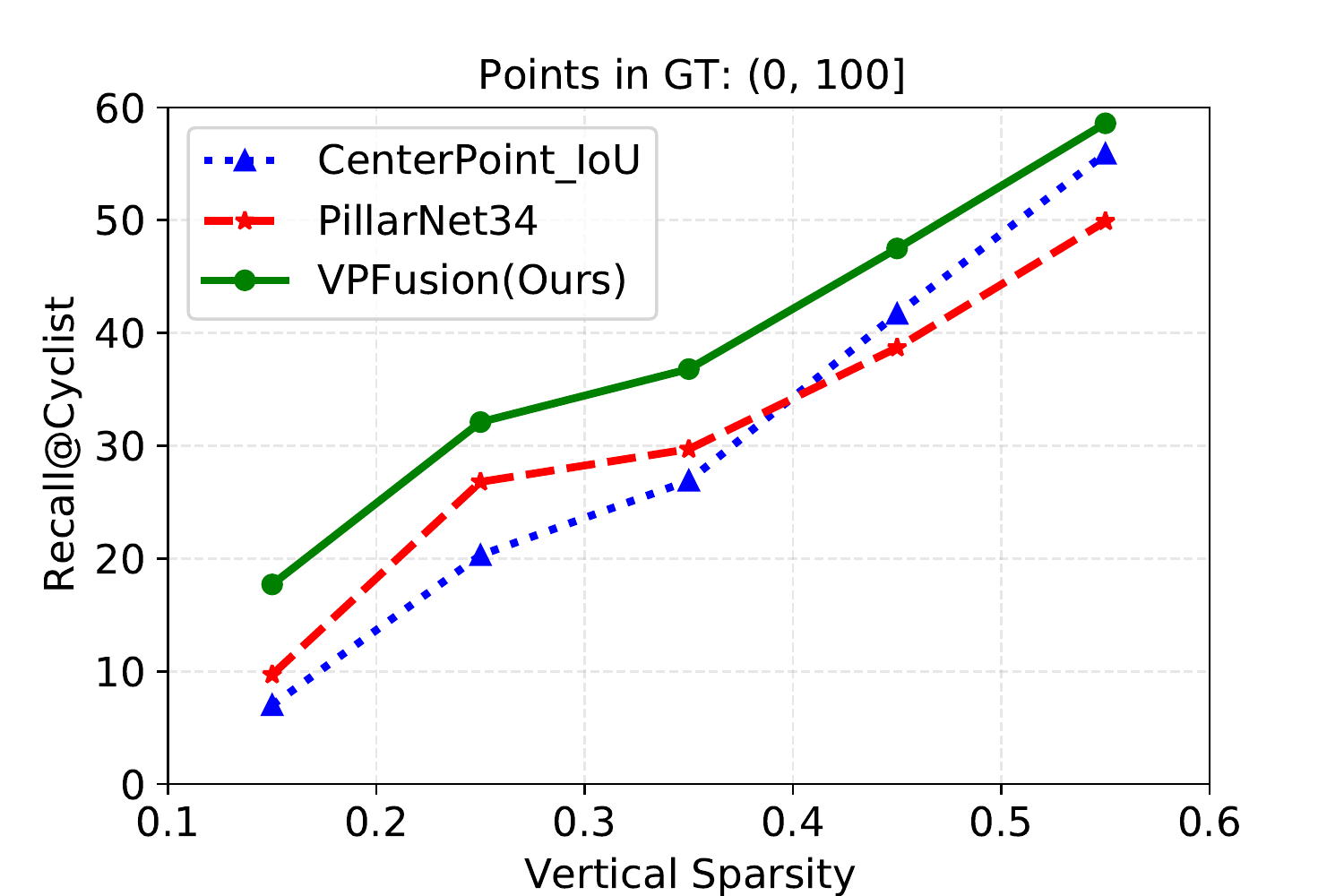}
    }
    \end{minipage}
    \begin{minipage}[t]{0.49\linewidth}
    \centerline{
        \includegraphics[width=\linewidth,trim=14 0 33 18,clip]{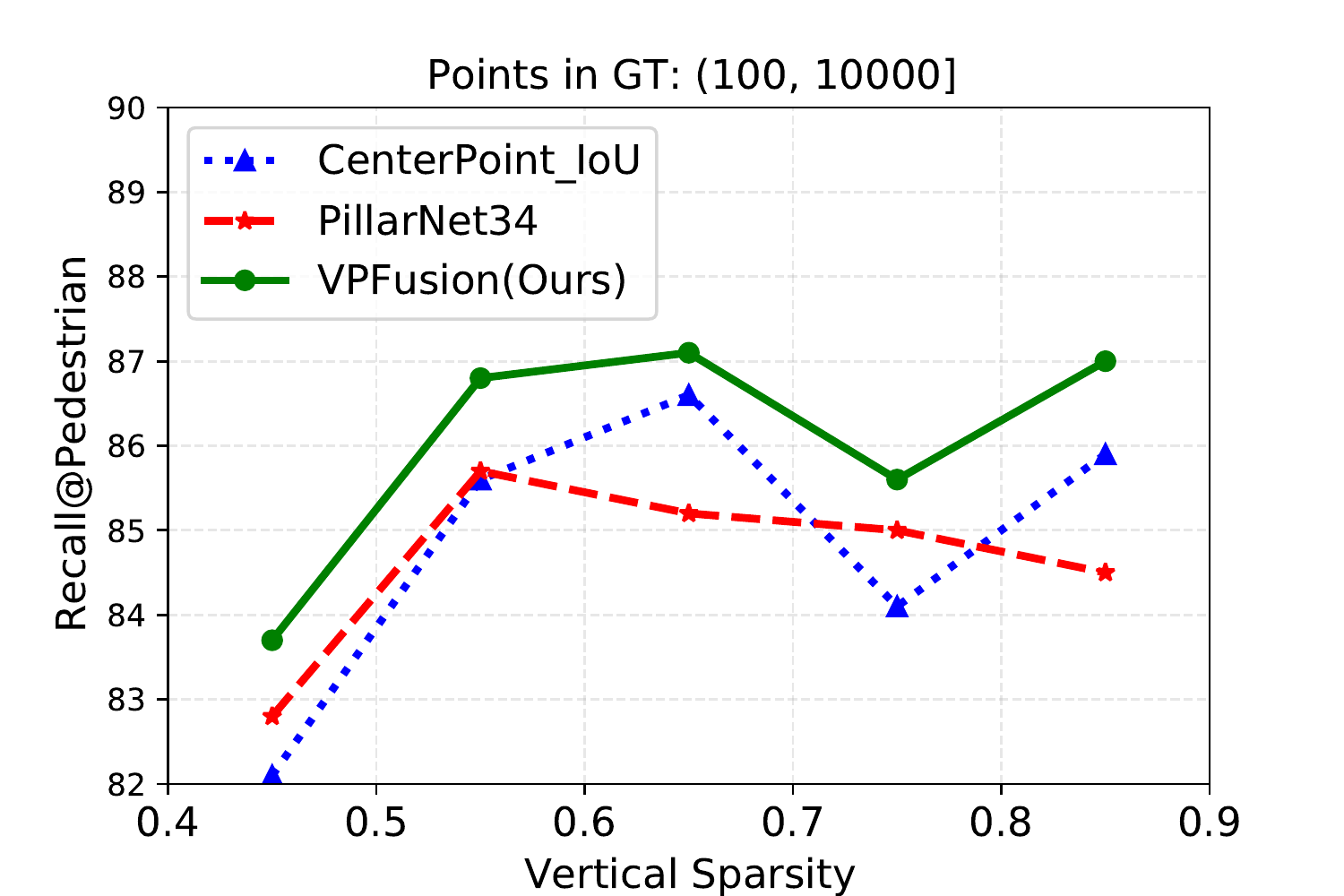}
    }
    \end{minipage}
}
\caption{Recall vs. Vertical Density comparison under different conditions. Our method possesses superior performance in varying situations and vertical densities.}
\label{fig:more_vpf_statistics}
\end{figure}

We delve deeper into the distinctions between voxels and pillars in Fig.~\ref{fig:more_vpf_statistics}. Specifically, to eliminate the influences from the horizontal distribution of points and the number of points in GT, we further split objects based on points count in GT and their horizontal occupancy. The horizontal occupancy is quantified as the geometric mean of the density along the X and Y axes, formulated as $\sqrt{S_{X} S_{Y}}$. For comparison, we use the CenterPoint\_IoU, PillarNet, and our VPF$_\mathrm{de}$ as a control group to underscore the differences between the conventional voxel/pillar-based encoding paradigm and our hybrid representation. As shown in Fig.~\ref{fig:more_vpf_statistics}, by synergistically combining the strengths of both voxels and pillars, our approach delivers consistent improvement across various conditions.

\subsection{Performance on nuScenes {\em val} set}

We presnet the comparison for the nuScenes {\em val} set in Tab.~\ref{tab:nuscenes_val}. VPF surpasses previous single voxel/pillar-based methods by a large margin, especially in NDS.

\begin{table}[h!]
    \centering
    \small
    \resizebox{0.81\linewidth}{!}{
    \begin{tabular}{c|c| c c c}
    \hline
    Table \#3 & Voxel\_size & mAP & NDS \\
    \hline
    CenterPoint & \multirow{4}{*}{0.75cm} & 59.22 & 66.48 \\
    PillarNet &  & 59.90 & 67.39 \\
    VoxelNeXt &  & 60.53 & 66.65 \\
    VPF$_\mathrm{de}$ &  & \textbf{61.83} & \textbf{68.89} \\
    VPF$_\mathrm{de}^{\dagger}$ &  & \textbf{64.68} & \textbf{71.05} \\
    \hline
    \end{tabular}}
    \caption{Non-ensemble comparison on the nuScenes {\em val} set. $\;^{\dagger}$: Use the fade strategy.}
    \label{tab:nuscenes_val}
\end{table}

\subsection{More Ablations}

\subsubsection{VPF$_\mathrm{sp}$.} We evaluate the impact of individual components in VPF$_\mathrm{sp}$ within Tab.~\ref{tab:vpf_sp_ablation}. Note that both vanilla VoxelNeXt and VoxelNeXt2D are applied as voxel/pillar-based baselines, and all experiments utilized the same detection head. As observed in the $3^{rd}$ row, the dual-branch voxel-pillar encoder, which uses element-wise summation for sparse voxel and pillar feature fusion, yields improvements of $1.9$ and $3.7$ mAPH compared to single representation methods. Additionally, when equipped with the SFL, VPF$_\mathrm{sp}$ significantly exceeds the performance of the baselines.

\subsubsection{Dense Fusion Neck.} In Tab.\ref{tab:dfn_branches}, we analyze the effects of DFN. We first study whether applying the single voxel or pillar branch as the dense neck could achieve similar performance. To this end, we retain either the single voxel or pillar branch in the final SFL, and apply the dense neck on the isolated dense voxel or pillar feature map as~\cite{yin2021center,shi2022pillarnet}. The first three rows of Tab.~\ref{tab:dfn_branches} validate the proposed fusion neck as the preferred choice. Moreover, we explore various fusion strategies in $3^{rd}$-$5^{th}$ rows, including element-wise summation, concatenation, and gated fusion~\cite{yoo20203d}. Despite similar performance outcomes, we apply the summation owing to its simplicity.

\begin{table}[t!]
    \centering
    \scriptsize
    \resizebox{\linewidth}{!}{
    \begin{tabular}{c| l | c c c c}
    \hline
    \multirow{2}{*}{Exp.} & \multicolumn{1}{c|}{\multirow{2}{*}{Settings}} & \multicolumn{4}{c}{LEVEL 2 APH} \\
    & & Veh. & Ped. & Cyc. & mAPH \\
    \hline
    1 & voxel & 67.6 & 67.2 & 69.8 & 68.2 \\
    2 & pillar & 67.5 & 63.7 & 68.1 & 66.4 \\
    3 & w/o SFL & 69.5 & 68.5 & 72.3 & 70.1 \\
    4 & with SFL & 70.5 & 69.8 & 72.4 & 70.9 \\
    \hline
    \end{tabular}}
    \caption{Ablation of VPF$_\mathrm{sp}$.}
    \label{tab:vpf_sp_ablation}
  \end{table}

\begin{table}[t!]
    \centering
    \scriptsize
    \resizebox{\linewidth}{!}{
    \begin{tabular}{c| l | c c c c}
    \hline
    \multirow{2}{*}{Exp.} & \multicolumn{1}{c|}{\multirow{2}{*}{Settings}} & \multicolumn{4}{c}{LEVEL 2 APH} \\
    & & Veh. & Ped. & Cyc. & mAPH \\
    \hline
    1 & Voxel-only & 69.3 & 65.4 & 68.4 & 67.7 \\
    2 & Pillar-only & 69.7 & 67.4 & 70.3 & 69.1 \\
    3 & Summation & 70.2 & \textbf{68.2} & \textbf{70.7} & \textbf{69.7}\\
    4 & Concatenate & 70.1 & 67.6 & \textbf{71.1} & 69.6 \\
    5 & Adaptive~\shortcite{yoo20203d} & \textbf{70.3} & 67.4 & 69.8 & 69.2 \\
    \hline
    \end{tabular}}
    \caption{Ablation of DFN. Parameter scales are comparable in Voxel-only and Pillar-only ablations.}
    \label{tab:dfn_branches}
  \end{table}

\section{Qualitative Results}

\begin{figure*}[t]
    \centerline{
        \begin{minipage}[t]{0.94\linewidth}
        \centerline{
            \includegraphics[width=\linewidth]{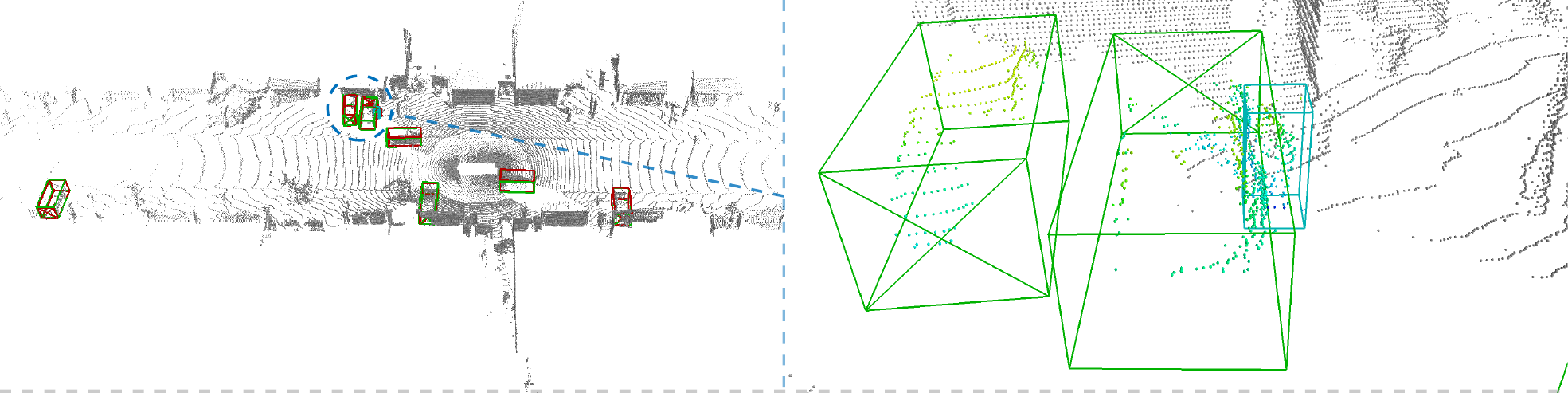}
        }
        \end{minipage}
    }
    \vspace{1pt}
    \centerline{
        \begin{minipage}[t]{0.94\linewidth}
        \centerline{
            \includegraphics[width=\linewidth]{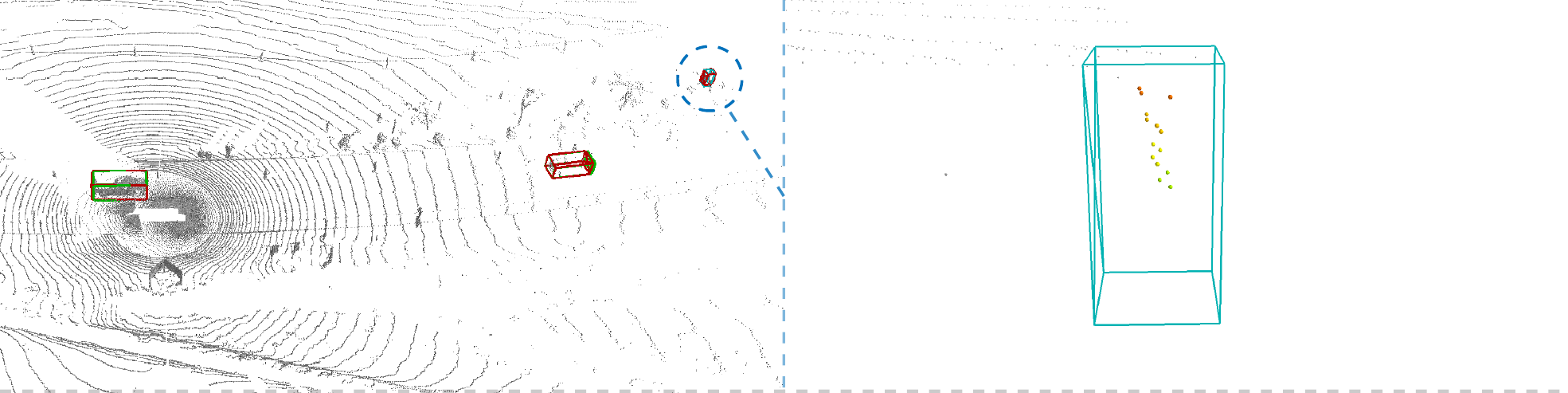}
        }
        \end{minipage}
    }
    \vspace{1pt}
    \centerline{
        \begin{minipage}[t]{0.94\linewidth}
        \centerline{
            \includegraphics[width=\linewidth]{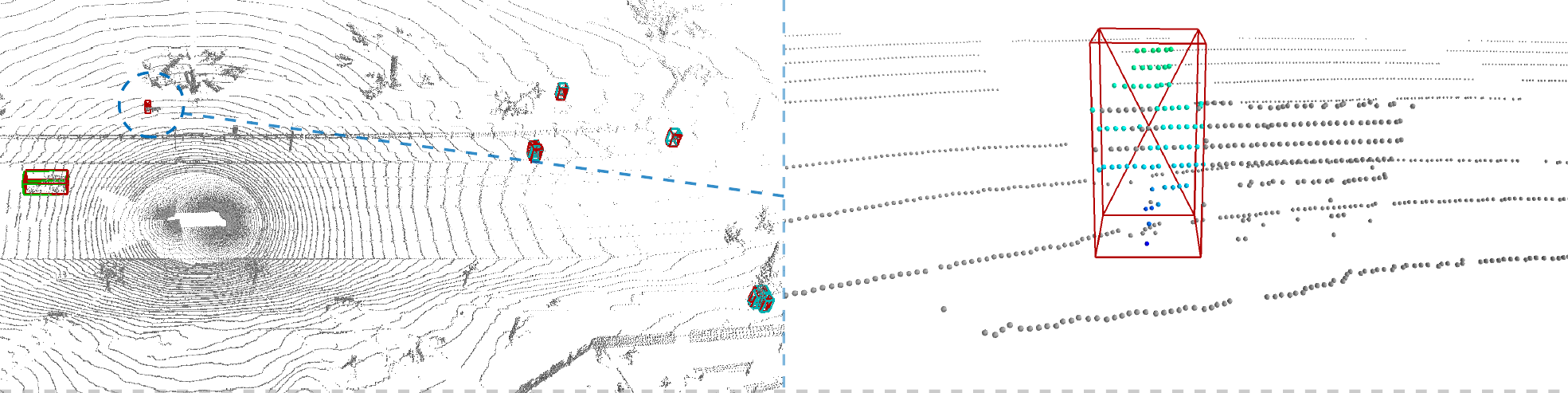}
        }
        \end{minipage}
    }  
    \vspace{1pt}
    \centerline{
        \begin{minipage}[t]{0.94\linewidth}
        \centerline{
            \includegraphics[width=\linewidth]{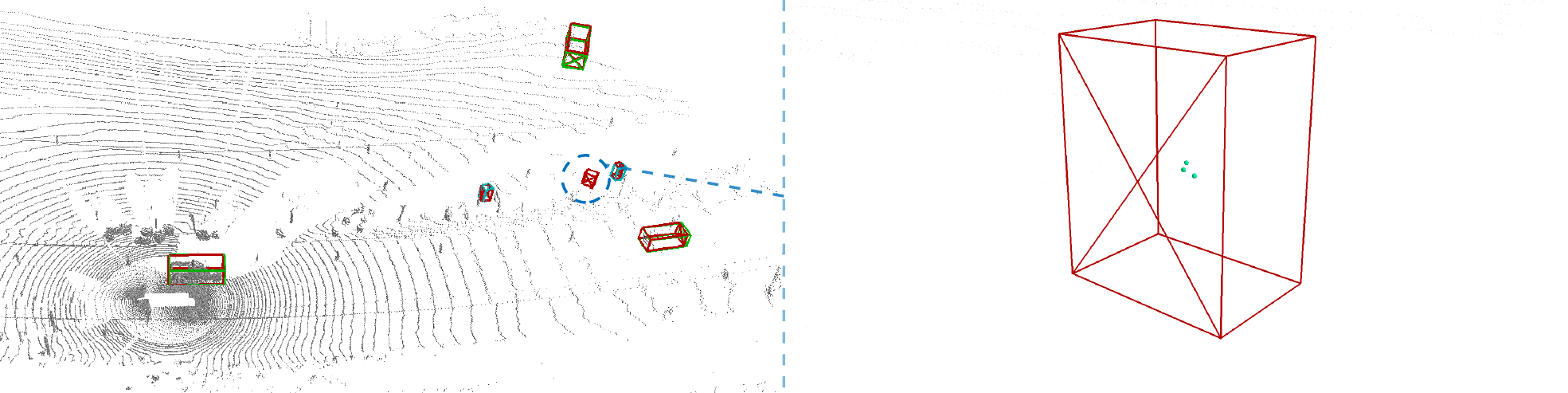}
        }
        \end{minipage}
    }  
    \caption{Visualization on the Waymo Open Dataset {\em val} set. Predicted bounding boxes for Vehicles, Pedestrians, and Cyclists are represented in green, cyan, and blue, respectively; ground truths are highlighted in red. The whole scene is displayed in the first column, while specific localized instances are detailed in the second column.}
    \label{fig:visualization}
\end{figure*}

In Fig.~\ref{fig:visualization}, we demonstrate some qualitative results. 
The first two rows demonstrate VPF can accurately recognize the objects that are tightly distributed or partially occluded. We also show some failure cases in the last two rows, which displays the situation of foreground and background points intertwine or points are extremely sparse. To our understanding, integrating texture details from images or accumulating past frames could help address these scenarios.

\end{document}